\documentclass[journal,hideappendix]{vgtc}      % final (journal style) without appendices
\onlineid{1357}

\vgtccategory{Research}

\vgtcpapertype{system}

\title{RingGesture: A Ring-Based Mid-Air Gesture Typing System Powered by a Deep-Learning Word Prediction Framework }
\author{%
  Junxiao Shen, Roger Boldu, Arpit Kalla, Michael Glueck, Hemant Bhaskar Surale, and Amy Karlson
}

\authorfooter{
  \item
  	Junxiao Shen is with Reality Labs Research, Meta and University of Bristol.
  \item
        Roger Boldu, Arpit Kalla, Michael Glueck, Hemant Bhaskar Surale, and Amy Karlson are with Reality Labs Research, Meta.
}

\abstract{%
Text entry is a critical capability for any modern computing experience, with lightweight augmented reality (AR) glasses being no exception. Designed for all-day wearability, a limitation of lightweight AR glass is the restriction to the inclusion of multiple cameras for extensive field of view in hand tracking. This constraint underscores the need for an additional input device. We propose a system to address this gap: a ring-based mid-air gesture typing technique, \textit{RingGesture}, utilizing electrodes to mark the start and end of gesture trajectories and inertial measurement units (IMU) sensors for hand tracking. This method offers an intuitive experience similar to raycast-based mid-air gesture typing found in VR headsets, allowing for a seamless translation of hand movements into cursor navigation. To enhance both accuracy and input speed, we propose a novel deep-learning word prediction framework, \textit{Score Fusion}, comprised of three key components: a) a word-gesture decoding model, b) a spatial spelling correction model, and c) a lightweight contextual language model. In contrast, this framework fuses the scores from the three models to predict the most likely words with higher precision. We conduct comparative and longitudinal studies to demonstrate two key findings: firstly, the overall effectiveness of \textit{RingGesture}, which achieves an average text entry speed of 27.3 words per minute (WPM) and a peak performance of 47.9 WPM. Secondly, we highlight the superior performance of the \textit{Score Fusion} framework, which offers a 28.2\% improvement in uncorrected Character Error Rate over a conventional word prediction framework, \textit{Naive Correction}, leading to a 55.2\% improvement in text entry speed for \textit{RingGesture}. Additionally, \textit{RingGesture} received a System Usability Score of 83 signifying its excellent usability.

}


\teaser{
  \centering
  \includegraphics[width=\linewidth]{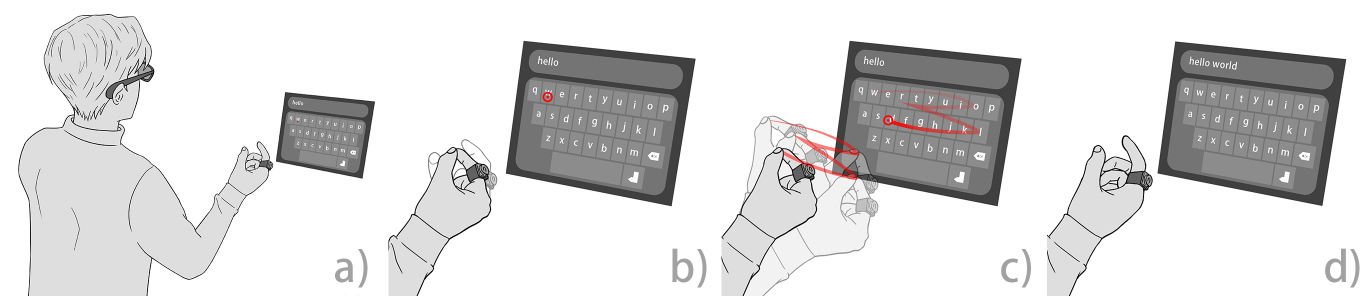}
  \caption{\textit{RingGesture}, a ring-based mid-air gesture typing system, enables users to input text both quickly and accurately. 
  The process unfolds as follows:
  a) The process begins when the user articulates their wrist, positioning the cursor over the initial letter of the desired word.
  b) Then, the user performs a pinch gesture with their thumb and index finger, marking the start of the cursor's trajectory.
  c) Subsequently, the user gestures the word's trajectory in mid-air to complete the input by articulating their wrist.
  d) Upon releasing the pinch, the deep-learning word prediction framework, \textit{Score Fusion}, predicts Top-K words, with the Top-1 word being pre-selected.
  }
  \label{fig:teaser}
}

\renewcommand{\manuscriptnotetxt}{}

\graphicspath{{figs/}{figs/}{pictures/}{images/}{./}} % where to search for the images

\usepackage{tabu}                      % only used for the table example
\usepackage{booktabs}                  % only used for the table example
\usepackage{lipsum}                    % used to generate placeholder text
\usepackage{mwe}                       % used to generate placeholder figures

\usepackage{mathptmx}                  % use matching math font

\usepackage{microtype}                 % use micro-typography (slightly more compact, better to read)
\PassOptionsToPackage{warn}{textcomp}  % to address font issues with \textrightarrow
\usepackage{textcomp}                  % use better special symbols
\usepackage{mathptmx}                  % use matching math font
\usepackage{times}                     % we use Times as the main font
\usepackage{cite}                      % needed to automatically sort the references
\usepackage{tabu}                      % only used for the table example
\usepackage{booktabs}                  % only used for the table example
\usepackage{amsmath}
\usepackage{multirow}
\usepackage{url}
\usepackage{floatrow}

\usepackage{booktabs}
\usepackage{array}
\usepackage{subcaption}
\usepackage{float}
\usepackage{dblfloatfix}

\usepackage{algorithm}
\usepackage[noend]{algpseudocode}

\newenvironment{s_itemize}{
\begin{itemize}[leftmargin=*]
  \setlength{\itemsep}{3pt}
  \setlength{\parskip}{0pt}
  \setlength{\parsep}{0pt}
}{\end{itemize}}

\newenvironment{s_enumerate}{
\begin{enumerate}[wide, labelwidth=!, labelindent=0pt]
  \setlength{\itemsep}{2pt}
  \setlength{\parskip}{0pt}
  \setlength{\parsep}{0pt}
}{\end{enumerate}}

\usepackage{array}
\usepackage{enumitem}
\usepackage{tabularx}
\graphicspath{{./figs/}}
\usepackage{enumitem}
\usepackage{float}
\usepackage{dblfloatfix} 

\newcommand{\changetext}[1]{\textcolor{black}{#1}}

\begin{document}

%%%%%%%%%%%%%%%%%%%%%%%%%%%%%%%%%%%%%%%%%%%%%%%%%%%%%%%%%%%%%%%%
%%%%%%%%%%%%%%%%%%%%%% START OF THE PAPER %%%%%%%%%%%%%%%%%%%%%%
%%%%%%%%%%%%%%%%%%%%%%%%%%%%%%%%%%%%%%%%%%%%%%%%%%%%%%%%%%%%%%%%

\firstsection{Introduction}

\maketitle

\begin{table*}[t]
\scriptsize
\centering
\caption{Summary of previous studies on one-handed text entry methods for QWERTY keyboards, with entry rates near or above 10 WPM. 
\textit{Directly Mapped Cursor} maps hand, head, arm or finger movements to cursor movement with direct projection.
In contrast, \textit{Indirectly Mapped Cursor} involves an intermediate algorithm between the physical movement and cursor movement. Studies involving \textit{Indirectly Mapped Cursor} typically demonstrate text entry rate under 10 WPM.
Note that we only report the text entry rate from novice users, as the experimental setup varies among users when measuring expert performance.}
\begin{tabular}{llllc}
\hline
Study & Interaction & Device Setup & Typing method (QWERTY) & Entry Rate (WPM) \\ \hline
% Reyal et al. \cite{reyal2015performance} & \textit{Direct Touch} & Smartphone Screen&  Gesture Typing & 27.0 \\
Markussen et al. \cite{markussen2014vulture} & \textit{Directly Mapped Cursor} - Hand & OptiTrack Hand Tracking&  Gesture Typing & 20.6 \\
Yu et al. \cite{yu2017tap} & \textit{Directly Mapped Cursor} - Head & Headset IMU Orientation &  Gesture Typing & 19.0 \\
% Kern et al. \cite{kern2023text} & \textit{Direct Touch} & Controller Tracking&  Gesture Typing & 17.5 \\
Zhao et al. \cite{zhao2023gaze} & \textit{Directly Mapped Cursor} - Arm & Wristband IMU Tracking&  Gesture Typing & 16.4 \\
Gu et al. \cite{gu2020qwertyring} & \textit{Direct Touch} & Ring IMU Tracking&  Gesture Typing & 13.8 \\
% Xu et al. \cite{xu2019pointing} & \textit{Directly Mapped Cursor} & Controller Tracking&  Gesture Typing & 13.7 \\
Henderson et al. \cite{henderson2020stat} & \textit{Directly Mapped Cursor} - Finger & Smartphone Screen &  Gesture Typing & 13.2 \\
Xu et al. \cite{xu2019tiptext} & \textit{Direct Touch} & On-Fingertip Sensors &  Touch Typing & 11.9 \\
Wang et al. \cite{Wang2015palm} & \textit{Direct Touch} & Vicon 3D Hand Tracking &  Touch Typing & 10.0 \\
Chen et al. \cite{chen2023DRG} & \textit{Direct Touch} & Ring IMU Tracking &  Gesture Typing & 9.9 \\
% Gong et al. \cite{gong2018wristext} & \textit{Indirectly Mapped Cursor} & On-Wristwatch Sensors &  Touch Typing & 9.9 \\
Gupta et al. \cite{gupta2019rotoswype} & \textit{Indirectly Mapped Cursor} & Ring IMU Tracking &  Gesture Typing & 9.2 \\ 
\hline
\end{tabular}
\label{tab:entry_rate_ranking}
\end{table*}

This paper focuses on one-handed text entry methods, necessitated by the occasional unavailability of both hands~\cite{chen2014swipeboard,jiang2022pinchtext, grossman2015typing, zhao2023gaze}, for lightweight augmented reality (AR) glasses in contrast to fully-fledged AR headsets (such as Apple Vision Pro~\cite{AppleVisionPro2023,AppleVisionProWikipedia}, Quest Series~\cite{OculusQuestSeries}, and HoloLens Series~\cite{MicrosoftHoloLens2,HoloLens2Wikipedia}).
We have analyzed prior research on one-handed text entry and found that most existing methods struggle with a range of limitations. 
These limitations include learnability challenges, lower performance ceiling, and intricate device setup. 
Learnability challenges stems from an indirect correlation between hand movement and keyboard key selection, and the introduction of numerous new keyboard layouts (with new key arrangements)~\cite{jiang2019hifinger, lee2019hibey, gong2018wristext, gupta2019rotoswype, grossman2015typing, rakhmetulla2020swipering, gu2020qwertyring}. 
Additionally, these methods typically demonstrate low entry speeds below 15 words per minute (WPM).
Some other methods also require intricate device setup procedures that require specific auxiliary devices like capacitive sensors on the fingertips~\cite{xu2019tiptext, peshock2014argot}.
Our observations highlight that gesture typing with the cursor directly mapped from body movements tends to yield the highest text entry rates, as evidenced by the systems developed by Markussen et al.~\cite{Vulture2014Markussen} using hand control (20.6 WPM), Yu et al.~\cite{yu_tap_2017} using head control (19.0 WPM), and Zhao et al.~\cite{zhao2023gaze} using arm control (16.4 WPM).
This efficiency can be attributed to the simplicity and intuitiveness of direct cursor projection combined with the rapid text entry facilitated by gesture typing.
Among these, Vulture~\cite{markussen2014vulture} methods offer the highest text entry rates, and are theoretically more ergonomic and less fatiguing when compared to using head and arm. 
However, they utilized OptiTrack~\cite{furtado2019opt} for hand tracking, an outside-in tracking method that comes with inherent deployment challenges. 
It requires the instrumentation of the user's entire arm to track movements of the arm, wrist, and fingers, and may at times lose tracking due to marker occlusion.

Consequently, we leveraged a ring device proposed by Kienzle et al.~\cite{kienzle2021electroring} to track hand positioning. 
This ring can track hand movements using its built-in IMU and detect stateful pinch actions using integrated electrodes~\cite{kienzle2021electroring}.
To this end, we refined our design space to concentrate on evaluating word-level gesture typing versus phrase-level gesture typing, the latter being less explored~\cite{xu2022phrase}.
We conducted a comparative study with 32 participants. Results showed no significant text entry performance difference between the two methods, but a preference for word-level typing emerged. 
We also tackled the Heisenberg Effect challenge, associated with input cross modality~\cite{Wolf2020Understanding} while performing mid-air pointing when with the discrete pinch actions, by introducing a customized filter-based algorithm.
Even still, IMU-based tracking inevitably introduces noise and drift into the cursor's trajectory, leading to inaccurate gesture typing decoding.
To guarantee swift and precise gesture typing, we proposed a novel deep-learning word prediction framework, \textit{Score Fusion}, which predict user's indented words based on not only user's gestured trajectories but also keyboard spatial information and contextual information from previous conversations. 
This framework integrated three crucial components through fusing the probabilistic scores of word candidates from the components: 1) a word-gesture decoding model; 2) a spatial spelling correction model; and 3) a lightweight contextual language model.

We conducted the second user study involving 16 participants with two primary objectives: 1) to evaluate the effectiveness of the \textit{RingGesture} system, and 2) to compare the proposed \textit{Score Fusion} algorithm with a conventional word prediction baseline, \textit{Naive Correction}, from \cite{shen2023fast}.
The results indicated that, firstly, \textit{RingGesture} achieves an average text entry rate of 27.3 WPM and a peak performance of 47.9 WPM. 
These results are comparable to mobile phone gesture typing performances, which are near 30 WPM \cite{reyal2015performance}. 
Secondly, we highlight the superior performance of the \textit{Score Fusion} framework, which offers a 55.2\% improvement in text entry speed over \textit{Naive Correction} (only achieving 17.6 WPM), due to improved word prediction from the \textit{Score Fusion}. 
This underscores the significance of the \textit{Score Fusion} framework for enabling a fast \textit{RingGesture} system.
Additionally, \textit{RingGesture} received a System Usability Score of 83, signifying its excellent usability.

In conclusion, our contributions are threefold:
\begin{enumerate}
    \item We propose a fast, accurate and easy-to-learn ring-based mid-air gesture typing system, \textit{RingGesture}, which enables users to perform text entry at rates (average entry rate: 27.3 WPM, novice entry rate: 26.4, expert entry rate: 32.5 WPM) comparable to mobile phone gesture typing rate.  
    \item We propose a novel deep-learning word prediction framework, \textit{Score Fusion}, which includes a word-gesture decoding model enabled by a novel data transformation process, a spatial spelling correction model enabled by a novel keyboard-layout-aware edit distance, and a novel pre-trained contextual language model while still being lightweight. 
    % This framework enhances \textit{RingGesture}'s text entry performance by 55\% for speed, compared to a conventional word prediction framework baseline, \textit{Naive Correction}.
    \item We conducted two studies to understand the value of our design decisions, the \textit{RingGesture} system, and the \textit{Score Fusion} framework: Study 1 to explore word-level gesture typing versus phrase-level gesture typing under \textit{Directly Mapped Cursor} interaction mode; Study 2 to demonstrate the efficiency of \textit{RingGesture}, and underscore the significant improvement of the \textit{Score Fusion} framework over a conventional word prediction baseline \textit{Naive Correction} (28.2\% improvement in uncorrected Character Error Rate, leading to 55.2\% improvement in text entry speed).
\end{enumerate}

\begin{figure}[t]
    \centering
    \includegraphics[width=0.45\textwidth]{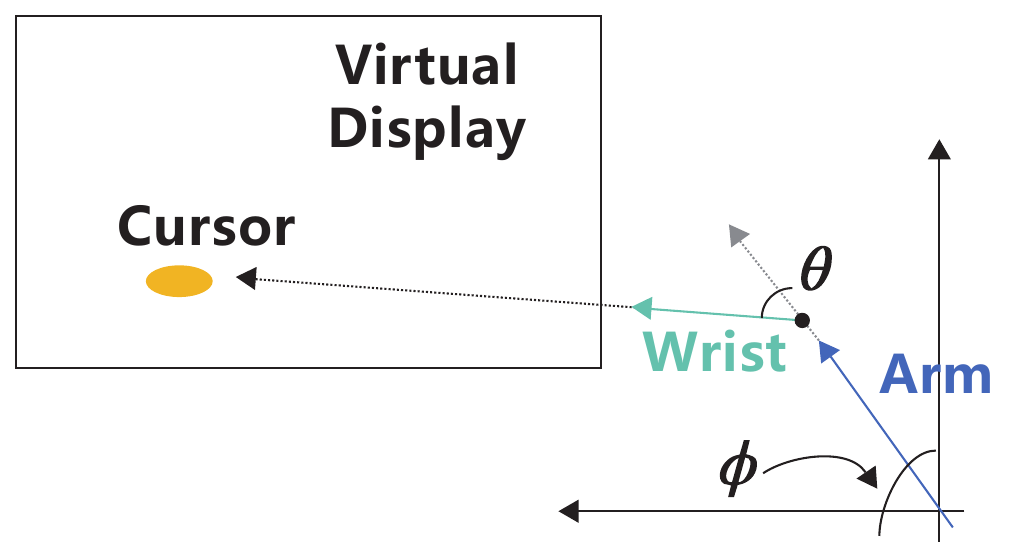}
    \caption{2D cursor control from arm and wrist. The cursor controlled by a ring is a direct mapping from $\theta$, and the cursor controlled by a wristband is a direct mapping from $\phi$.}
    \label{fig:imu_cursor}
\end{figure}

\section{Related Work}

Various studies of text entry methods in AR/VR suggest typical entry rates of 5 to 26 WPM~\cite{xu2019pointing,jdVI,shen2022personalization,Evaluating2023Dudley, shen2023fast}. It is evident that leveraging users' existing typing skills with QWERTY keyboards and simple interactions provides the best performance~\cite{kim2023star,Dudley2019virtual,dudley2018fast}, while abstract mappings and complex designs hinder text entry rate~\cite{lee2017vitty,Speicher2018selection,Ogitani2018handcamera,jiang2019hifinger,lepouras2018comparing,YU2018pizzatext}.

\subsection{One-Handed Text Entry}

This section critically reviews and compares notable one-handed text entry research. 
Our focus is narrowed to those methods that have achieved performance rates near or surpassing 10 WPM and utilize QWERTY keyboards. 
Table~\ref{tab:entry_rate_ranking} presents a comparative overview of these one-handed text entry studies. 

Markussen et al.~\cite{markussen2014vulture} proposed Vulture which is a mid-air gesture typing keyboard, and utilized OptiTrack hand tracking for \textit{Directly Mapped Cursor} control via hand movements, achieving an entry rate of 20.6 wpm. This method exemplifies high efficiency in text entry by leveraging intuitive hand movements, closely mirroring physical interactions in the real world. The direct manipulation facilitated by this approach suggests a significant potential for enhancing user experience in AR and VR environments through natural interaction paradigms. 
\changetext{Please note the term \textit{Directly Mapped Cursor} does not refer to the fingertip directly touching the keyboard. Instead, \textit{Directly Mapped Cursor} refers to the direct mapping between cursor movement and finger movement, rather than direct physical contact.}

Zhao et al.~\cite{zhao2023gaze} also employed a similar \textit{Directly Mapped Cursor} control-based mid-air gesture typing technique, but utilized arm movements instead, incorporating a wristband with in-built IMU to track the arm's position. 
They achieved an entry speed of 16.4 WPM. 
Despite introducing the `Speedup' method, which accelerates the cursor towards the user's gaze fixation point to enhance text entry rate, the final improved speed reached only 17.1 WPM, which is still significantly lower than the 20.6 WPM achieved by Vulture~\cite{markussen2014vulture}.
One factor to this reduced speed is that using the arm for control introduces more extensive movement, inherently leading to slower speeds. 
Additionally, this method results in greater fatigue as more torque is needed for the arm (\(\alpha\)) as compared to the wrist (\(\phi \)) because the arm's greater length, which then demands more force for the same orientation change (see Figure~\ref{fig:imu_cursor}.).

Similarly, Yu et al.\cite{yu2017tap} introduced a technique based on head orientation control via a headset to navigate the cursor, with an entry rate of 19.0 wpm. This method diverges from hand-based interaction by utilizing head movements for text entry, presenting an alternative that, while effective, introduces a different set of ergonomic considerations and potential user adaptation challenges.

Another study closely related to our work is RotoSwype by Gupta et al.~\cite{gupta2019rotoswype}, which also explores mid-air gesture typing through a ring device. However, unlike our \textit{Directly Mapped Cursor} interaction, RotoSwype utilizes an \textit{Indirectly Mapped Cursor} mode, implementing an indirect mapping strategy where wrist rotations correspond to cursor movements. This method, however, presents considerable challenges in user adaptation due to its non-intuitive mapping system.
Gupta et al.~\cite{gupta2019rotoswype} noted, `Participants found it difficult at first to understand the mapping of angular movements to the flat pointer motion on-screen, especially for diagonal motion.' Consequently, the text entry speed for novices was recorded at 9.2 WPM, underscoring the inherent learning curve associated with this innovative interaction technique.

Additionally, Henderson et al.\cite{henderson2020stat} utilized smartphones for AR cursor interaction, achieving a 13.2 WPM rate, suggesting mobile tech integration can ease text entry access. Yet, this contradicts the goal of AR glasses reducing mobile phone dependence. Similarly, wearable tech for text entry was explored by Xu et al.\cite{xu2019tiptext}, Gu et al.\cite{gu2020qwertyring} and Chen et al~\cite{chen2023DRG}, with fingertip and ring-based sensors reaching 11.9 WPM, 13.8 WPM and 9.9 WPM, respectively. These methods present innovative, albeit less adaptable AR interaction solutions.

The top 3 text entry methods in Table~\ref{tab:entry_rate_ranking} are based on the \textit{Directly Mapped Cursor}-based mid-air gesture typing approach. 
However, each method has its drawbacks. Vulture~\cite{Vulture2014Markussen} relies on OptiTrack for hand tracking which is a `Wizard-of-Oz' technique, Yu et al.~\cite{yu_tap_2017}'s system uses head movements for cursor control, which introduces ergonomic challenges, and Zhao et al.~\cite{zhao2023gaze}'s system can cause arm fatigue due to the involvement of whole arm movements.
Our system, \textit{RingGesture}, improves upon Vulture~\cite{Vulture2014Markussen}'s mid-air gesture typing approach by using a ring equipped with Inertial Measurement Units (IMUs), allowing for cursor control through hand movements alone. 
Recognizing that IMUs may lead to noise and drift, resulting in inaccurate cursor control and subsequently inaccurate word-gesture decoding, we proposed a deep-learning word prediction framework, \textit{Score Fusion}, to enhance word prediction accuracy, thereby increasing the rate of text entry.

\begin{figure}[t]
    \centering
    \includegraphics[width=0.6\textwidth]{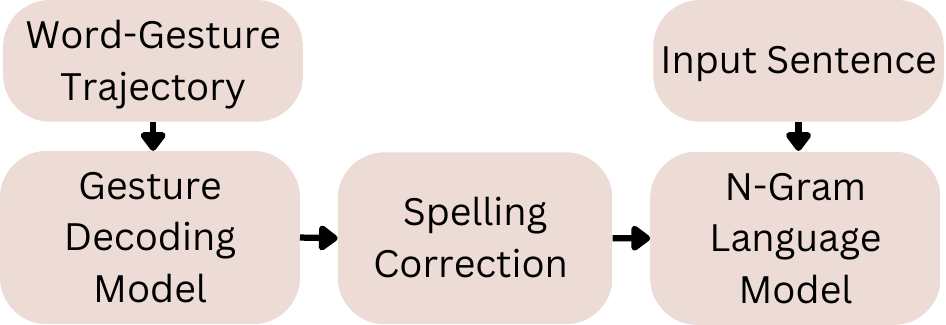}
    \caption{The conventional word prediction framework, \textit{Naive Correction}, operates in a sequential fashion: The predictions from the word-gesture decoding model are firstly corrected for misspellings by the edit-distance-based spelling correction model, which functions by calculating the edit distance with the word candidates in the corpus. The corrected candidates are then re-ranked by an N-Gram language model~\cite{jurafsky2019speech} based on the previously enter N-1 words.}
    \label{fig:sequential_correction}
\end{figure}

\subsection{Word Prediction in Text Entry}

\changetext{
Word prediction in text entry systems involves word decoding, spelling correction, and language modeling. The decoding model interprets the user's input patterns into raw predictions, forming a sequence of characters, but it may introduce spelling errors. The spelling correction model rectifies these inaccuracies, while language modeling further enhances word prediction by considering the context provided by previous words. Shen et al.~\cite{shen2023fast} connect these components sequentially, as illustrated in Figure~\ref{fig:sequential_correction}. The word prediction system, referred to as \textit{Naive Correction} in this paper, enhances accuracy through a sequential process involving edit-distance-based spelling correction and N-Gram language model re-ranking. Despite this system being the state-of-the-art word prediction framework for gesture typing, it has several drawbacks: it only considers limited contextual scope, ignoring global probabilities across the entire dataset. Additionally, it lacks deep semantic understanding, leading to potential inaccuracies in certain contexts.
We propose a novel word prediction system that integrates correction and language modeling through a probabilistic approach, allowing the system to consider global probabilities rather than processing them sequentially. Additionally, our approach enhances spelling correction by incorporating spatial information from the keyboard and advances the N-Gram language model by utilizing much longer contexts, while maintaining a lightweight design.
}

% \subsubsection{Word Decoding}

% \subsubsection{Spelling Correction}

General spelling correction encompasses a wide array of contexts, including document editing and processing digital texts in databases or on online platforms.
Its primary objective is to identify and correct errors throughout entire sequences of words~\cite{CaiRijke2016,BabaSuzuki2020,PirinenLinden2014}. 
In contrast, spelling correction in text entry systems on mobile devices is specifically tailored for real-time user inputs. 
It focuses on correcting errors in single words as they are typed, where the errors not only come from users but also from text decoders in a probabilistic text entry system~\cite{dudley2018fast}.
% Therefore, despite general spelling correction has significantly evolved~\cite{Hldek2020SurveyOA}, most of the advanced methods could not be applied to word error correction in text entry systems directly. 
Spelling correction in text entry systems commonly relies on the edit distance method, prized for its ease of integration with custom word corpora and modifications to achieve microsecond latency~\cite{Mitton1996, Kukich1992}. 
However, its accuracy for QWERTY-based text entry systems is limited~\cite{BabaSuzuki2020}. 
One of the major reasons is that it fails to consider the spatial information of the keyboard. 
This oversight leads to reduced correction accuracy by offering phonetically or orthographically similar but irrelevant corrections, and to inefficient error prediction due to the inability to accurately anticipate mistyped words based on common finger movements and miskeying patterns, thus diminishing its overall effectiveness. 
\changetext{
Our paper addresses this gap by proposing a novel spatial edit distance that incorporates the spatial information of a gesture typing keyboard. Additionally, we transformed the spatial edit distance into a probabilistic-based measure, allowing seamless integration with a probabilistic word decoder.
}
% Our paper addressed this gap by proposing a novel spatial edit distance that incorporates the spatial information of a gesture typing keyboard. Furthermore, we transformed the spatial edit distance to probablistic-based such that it could be seamleslly integrated to a probablistic word decoder. 

Language modeling represents another important approach for enhancing word error correction. 
N-gram language models are commonly utilized for meeting latency demands owing to their ease of implementation and explainability~\cite{Pauls2011FasterAS,jurafsky2019speech}. 
With the advancements in deep learning technologies, models based on deep learning have been progressively incorporated into language modeling~\cite{bengio2003neural, goldberg2017neural,sundermeyer2012lstm}. However, there has been limited research on contextual language modeling for text entry systems. 
Contextual information is a crucial element in text entry, which includes elements such as conversation history and other context tags like places, times, and hobbies of the users, etc.
Shen et al.\cite{shen2022kwickchat} proposed for the first time a contextual language model based on GPT-2 for Augmentative and Alternative Communication (AAC) use cases. However, the Generative Pre-trained Transformer-2 (GPT-2)\cite{Radford2019GPT2} model has significant latency when operated on a mobile device, while this latency is acceptable for the AAC use case in Shen et al.~\cite{shen2022kwickchat}.
% Moreover, their model's sentence prediction does not operate at the per-word input level, thus it does not have a strong constraint on latency.
\changetext{
% We propose a novel method that reinventing the contextual language model by training on Long Short Term Memory (LSTM) models instead of pre-training on transformers. 
We propose a novel method that transforms contextual language modeling through pre-training with Long Short Term Memory (LSTM) models instead of pre-training with transformers.
This results in a lightweight model architecture, with the final contextual language model being only 7 megabytes (MB) in size and capable of running in real-time on a mobile device.
}

% \changetext{

% }

\section{User Study 1: Word-Level versus Phrase-Level Gesture Typing}

We began by investigating whether the simple act of removing the delimitation requirements between words could accelerate mid-air gesture input under the \textit{Directly Mapped Cursor} interaction mode. 
\changetext{As we are proposing a novel and comprehensive text entry system that ranges from interaction design to backend architecture design, the choice between word-level and phrase-level typing is a fundamental component of the interaction design. Therefore, it is important to conduct a study to explore this aspect.}

While Xu et al.~\cite{xu2022phrase} have conducted studies comparing phrase-level to word-level gesture typing on smartphones, their approach relies on \textit{Direct Touch}. 
This differs from our \textit{Directly Mapped Cursor} mode.
Consequently, Xu et al.'s findings~\cite{xu2022phrase} may not be directly applicable to our context. 
Therefore, we conducted our own study using a touchpad to simulate \textit{Directly Mapped Cursor} interaction. Controlling a cursor by swiping on a touchpad directly translates fingertip movements into cursor movements on the screen. 
This method is particularly advantageous because it provides an accurate representation of our swipe path, serving as the ground truth. 
In contrast, most other methods, such as those based on inertial measurement units (IMUs) or camera tracking, introduce noise, detracting from the fidelity of tracking. 
Thus, using a touchpad enables us to simulate perfect tracking, which is crucial for the precision required in our comparisons of gesture typing at the phrase level and at the word level within this \textit{Directly Mapped Cursor} interaction mode.
The following are the details of the study:

\begin{s_enumerate}
    \item \textbf{Participants}: 
    We recruited 32 volunteers as participants through an internal mailing list, who had an average age of 33 (range 18-64, standard deviation 10.91).
    The group comprised 18 males, 13 females, and 1 participant who chose not to disclose their gender.
    27 participants are right-handed, and 5 participants are left-handed.
    \item \textbf{Apparaturs}: 
    Participants controlled a cursor using a Sensel Touchpad~\cite{sensel_haptic} placed on the table in front of a monitor, with the cursor displayed on a virtual keyboard on a monitor. The monitor was connected to a Lenovo PC (ThinkStation) equipped with an Intel Xeon processor.
    \item \textbf{Phrase Set}:
    The phrase set used was collected from two sources: the Enron Mobile Corpus~\cite{klimt2004enron} and the MacKenzie phrase set~\cite{mackenzie2003phrase}.
    This combined phrase corpus encompassed a total of 42,612 unique phrases. 
    Each phrase in this corpus exhibited an average length of 5.3 words, with a minimum length of 2 words and a maximum length of 7 words. 
    \item \textbf{Procedure}: 
    Participants in the study were directed to execute tasks under two distinct conditions: gesture typing at the word level and at the phrase level. 
    To ensure impartiality, these conditions were counterbalanced. 
    Each condition consisted of 40 phrases, selected uniformly from the aforementioned phrase set.
    Under word-level gesture typing, participants are instructed to delimit after swiping for each word by lifting up the finger from the touchpad.
    In contrast, under phrase-level gesture typing, participants are instructed to delimit only when the entire phrase is completed. 
    We implemented a pseudo-decoder, based on the model proposed by Shen et al.~\cite{shen_ismar2021}, that simulates an ideal decoder. 
    This decoder predicts correct words as long as at least 70\% of the gesture trajectory passes through the designated tolerance region for each character in the swiped word or phrase. To simulate the decoder's capability to manage ambiguous inputs effectively, the key region is defined to be four times larger than the actual size of the keys.
    Before starting each condition, participants were allowed to practice with 5 phrases. During the condition, participants could rest for up to 2 minutes after every 10 phrases. To advance to the next phrase, participants press the \texttt{Space Bar} button on the keyboard. 
    At the end of the study, participants were invited to fill out a post-study questionnaire. This included a Likert scale rating on various aspects for both word-level and phrase-level gesture typing: 1) \textit{Easy to Type}, 2) \textit{Easy to Learn}, 3) \textit{Fast to Type}, 4) \textit{Prediction is Accurate}, 5) \textit{Hand Feels Fatigued}, 6) \textit{Eyes Feel Fatigued}. 
    Additionally, participants were asked the following open-ended question: `Between word-level typing and phrase-level typing, did you find one method superior to the other? If so, which one and why?'
    Each condition for one participant took around 30 minutes to complete.
    \item \textbf{Evaluation Measures:}
    We report the results of the studies using the following metrics:
    \begin{s_itemize}
\item Words Per Minute (WPM) is represented mathematically as:
    \[
    WPM = \frac{\text{Total Words Typed}}{\text{Time in Minutes}}
    \]
    
    \item Character Error Rate (CER) can be quantified using the formula:
    \[
    \frac{\text{Minimum Number of Insertions, Deletions, and Substitutions}}{\text{Length of Stimulus Text}}
    \]
    
    \item Uncorrected Character Error Rate (Uncorrected CER) and Corrected Character Error Rate (Corrected CER) are defined for the predicted text output before and after correction interventions, respectively, with corrections including word deletions and re-entries, as follows:
    \[
    \text{Uncorrected CER} = \frac{\text{Number of Errors in Initial Prediction}}{\text{Length of Stimulus Text}}
    \]
    \[
    \text{Corrected CER} = \frac{\text{Number of Errors After Corrections}}{\text{Length of Stimulus Text}}
    \]
    \end{s_itemize}

\end{s_enumerate}

\begin{figure}[t]
    \centering
    \includegraphics[width=0.7\textwidth]{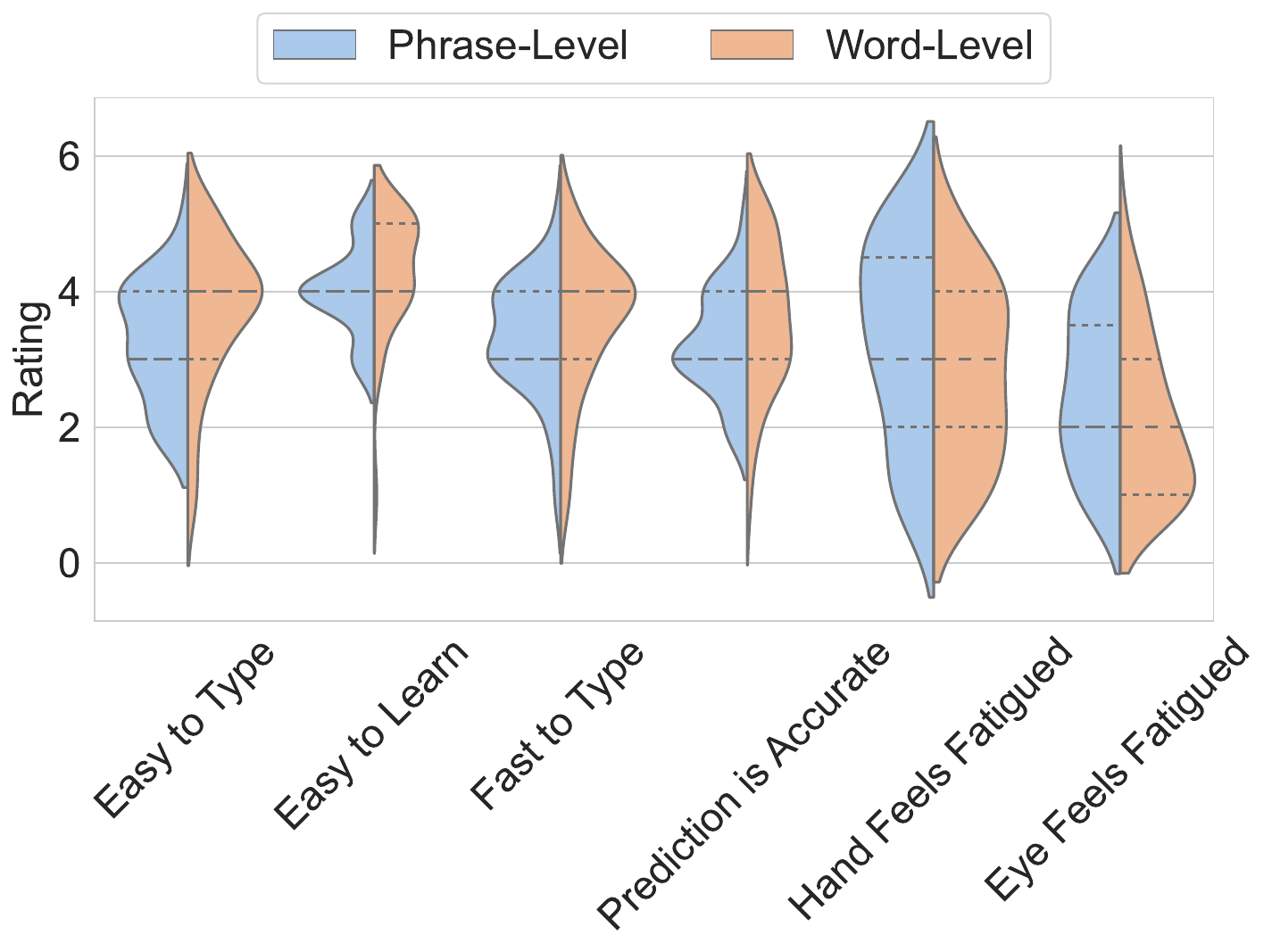}
    \caption{
    Violin plots of answers to subjective rating questions scored on 5-point Likert scales. Violin plots are modified box plots that add estimated kernel density plots to the summary statistics displayed by box plots. The 5-point Likert scales ranged from 1 (strongly disagree) to 5 (strongly agree).
    }
    \label{fig:study1_response}
\end{figure}

Initially, our analysis revealed no significant difference in text entry speed between word-level and phrase-level gesture typing, which were measured at 24.7 WPM and 25.5 WPM, respectively, accompanied by uncorrected character error rates of 7.8\% and 9.2\%.
To further explore the impact of typing conditions on text entry speed and accuracy, we performed an ANOVA analysis. The findings from this analysis indicated that there were no significant differences in both text entry speed (F= 2.14, p = 0.13) and accuracy (F = 1.89, p = 0.24) across the two conditions. This underscores the similarity in performance between word-level and phrase-level gesture typing in terms of both speed and precision.

Then we analyze the subjective feedback collected from the post-study questionnaire. 

\begin{s_itemize}
    \item \textit{Ratings on Six Aspects}:
    Figure~\ref{fig:study1_response} illustrates a comparison between phrase-level and word-level input methods across various aspects. Observing the shapes and distributions, Phrase-level input generally have a broader spread in ratings for ease of typing and learning, which indicate a more varied user experience. Word-level input, on the other hand, score consistently higher for being perceived as fast to type and having accurate predictions, with ratings clustered around higher medians, suggesting users may find it more efficient in these respects. When it comes to fatigue, both hand and eye fatigue are reportedly lower with word-level input, as reflected by the denser concentration of lower ratings. Overall, while there's some overlap in user responses, the data suggests a preference for word-level input in terms of speed, accuracy, and reduced fatigue.
    \item \textit{Preferences}:
    We further analyze the responses of participants regarding their preference between word-level typing and phrase-level typing.
    There were 19 participants who explicitly favored word-level typing. The reasons include: 1) \textbf{Familiarity} (6 mentions): Word-level typing is more similar to traditional typing methods, where each word is separated by `hitting the spacebar'. 2) \textbf{Cognitive Load} (5 mentions): Some users mentioned that their brains think in terms of words rather than phrases, making word-level typing more natural. 3) \textbf{Less Fatigue} (3 mentions): Some respondents indicated that word-level typing is less tiring because it doesn't require the user to hold and drag for long periods. 4) \textbf{Feedback} (2 mentions): Users get immediate feedback after typing each word, which helps them correct errors on the go. 5) \textbf{Coordination} (3 mentions): A few respondents found it easier to coordinate their eyes and hands while typing at the word level.
    However, 8 participants preferred phrase-level typing. The reasons include: 1) \textbf{Efficiency} (3 mentions): Phrase-level typing allows users to type longer sentences more quickly, as it eliminates the need to delimitate between words. 2) \textbf{Convenience} (3 mentions): Users found it convenient that phrase-level typing automatically segmented the words on their behalf. 3) \textbf{Accuracy} (2 mentions): Some users found phrase-level typing more accurate.
    Finally, 5 respondents did not express a clear preference for either word-level or phrase-level typing.
\end{s_itemize}

Given that phrase-level gesture typing did not yield any notable enhancements over word-level gesture typing as indicated, coupled with the fact that a larger user base preferred the latter, we opted for word-level gesture typing.

\section{RingGesture}
% \changetext{
% Study 1 investigated the interaction design of the \textit{RingGesture} system. This section provides a comprehensive overview of the backend model design of the RingGesture system. It includes the algorithm designed to overcome the Heisenberg Effect challenges associated with input cross-modality during mid-air pointing with discrete pinch actions, as described by Wolf (2020)~\cite{Wolf2020Understanding}. Additionally, it introduces our novel deep-learning word prediction framework, \textit{Score Fusion}, and discusses the details of its three individual components.
% }

\changetext{Study 1 investigated the interaction design of the \textit{RingGesture} system. This section provides a comprehensive overview of the backend architecture design of the \textit{RingGesture} system. It includes the algorithm designed to overcome the Heisenberg Effect challenges associated with input cross modality while performing mid-air pointing when with the discrete pinch actions~\cite{Wolf2020Understanding}, and our novel deep-learning word prediction framework, \textit{Score Fusion}.}

% In this section, we offer a comprehensive overview of the \textit{RingGesture} system, 

\subsection{Ring-Based Mid-Air Pointing and Selection}

We created a ring device with a reference design from ElectroRing proposed by Kienzle et al.~\cite{kienzle2021electroring}. This ring detects \textit{touch} and \textit{release} events of a pinch gesture by monitoring changes in an electrical signal. Furthermore, the ring uses an IMU for 2D cursor tracking by transforming accelerometer and gyroscope data into quaternions, converting these into polar coordinates, and then mapping them to Cartesian coordinates. The gain parameter for the control display is set to 1.8.
% We term the process of employing a ring for raycast interactions as \textit{RingCast} interaction.

While we effectively utilize the ElectroRing design~\cite{kienzle2021electroring} for pinch detection and IMU-based 2D cursor tracking, our system encountered a challenge in the context of mid-air gesture typing: Heisenberg Effect associated with input modality crosstalk~\cite{Wolf2020Understanding}. This issue arises when a discrete input like a pinch inadvertently alters the virtual cursor's position, resulting in an inaccurate selection point during mid-air pointing and selection interactions.

Therefore, to counteract the abrupt displacement introduced by pinch actions, we suggest a filter-based strategy. This approach dynamically determines the filtering level within an exponential smoothing filter by resolving the subsequent optimization problem:

\[
\alpha_0 = \underset{\alpha}{\mathrm{arg\,min}} \left\{ \lambda \sigma \underbrace{\sqrt{\frac{\alpha}{2 - \alpha}}}_{\text{Noise rejection}} + (1 - \lambda) \underbrace{\frac{(1 - \alpha)\Delta}{\alpha}}_{\text{Tracking error}} \right\},
\]

In this equation, $\sigma$ signifies the level of sensing noise, $\Delta$ provides an estimate of the signal's velocity, and $\lambda$ serves as a parameter that balances noise rejection (the left term in the minimization above) and infinite horizon tracking error in response to an input ramp (the right term in the minimization above). The $\lambda$ parameter is preset to $0.75$ with preliminary experiments. 

\begin{figure}[t]
    \centering
    \includegraphics[width=0.6\textwidth]{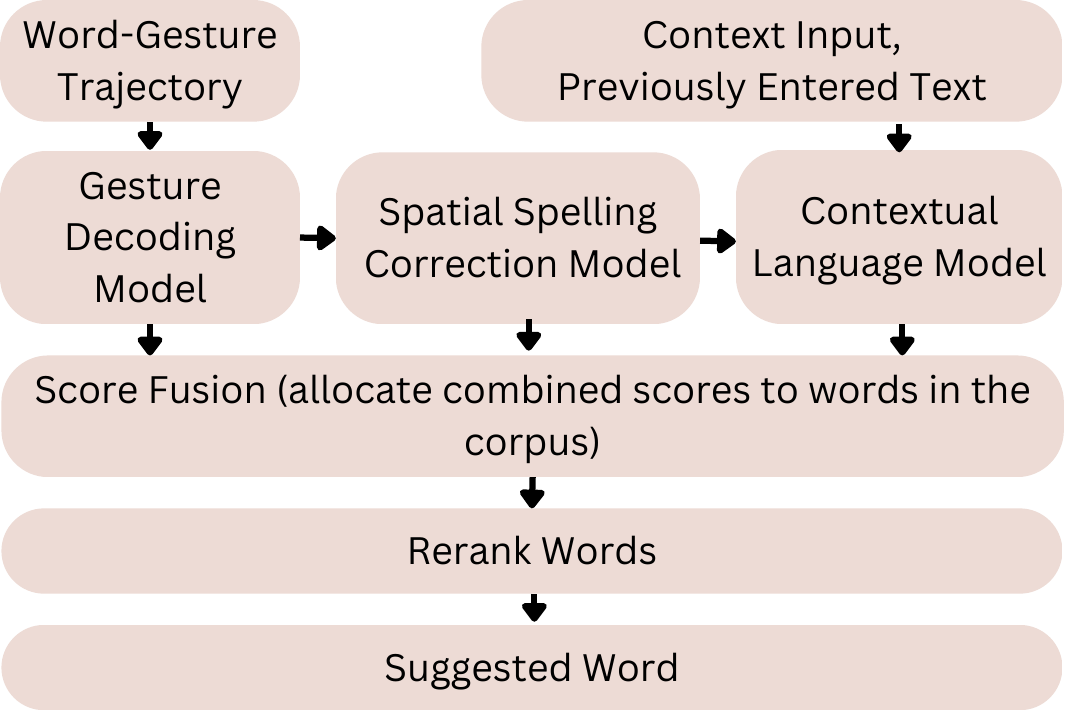}
    \caption{Our novel deep-learning word prediction framework, \textit{Score Fusion}, operates in an integrated fusion process: This fusion process evaluates each word suggestion by considering its initial decoding score, its likelihood of being a spatial spelling correction, and its contextual relevance. 
    The resulting blended score aims to ensure that the final suggestions are derived from an accurate word-gesture decoding model while also being enhanced for typographical precision, keyboard-layout-awareness, and contextual relevance.}
    \label{fig:score_fusion}
\end{figure}

\subsection{Score Fusion}
To mitigate the issues posed by input signal noise, such as hand jitter and IMU drift that lead to inaccuracies in word-gesture decoding, we have developed a deep-learning framework for word prediction, \textit{Score Fusion}.
This framework consists of three distinct components that compute the logarithmic probability of a word within a corpus based on a given word-gesture trajectory, as illustrated by Figure~\ref{fig:score_fusion}.
These individual scores are then consolidated to provide a composite score for the words across the corpus. Subsequently, we reorder these scores to present the highest-ranked words as the suggested options.

\subsubsection{Word-Gesture Decoding Model}
Deep-learning-based decoders~\cite{alsharif2015long, shen2023fast} have demonstrated significant advancements over traditional shape-matching-based decoders~\cite{kristensson2004shark2}. Motivated by these advancements, we aimed to train a deep-learning-based decoder tailored to our specific use case. However, deep learning models necessitate a substantial volume of training data to avoid overfitting. To address this challenge, we utilized a large-scale, publicly available gesture typing dataset, How We Swipe Dataset~\cite{leiva2021we}. Given that this dataset was collected from a different keyboard layout, we proposed a novel method to transform the trajectories to match our customized keyboard layout as follows:

\begin{figure}[t]
    \centering
    \includegraphics[width=\textwidth]{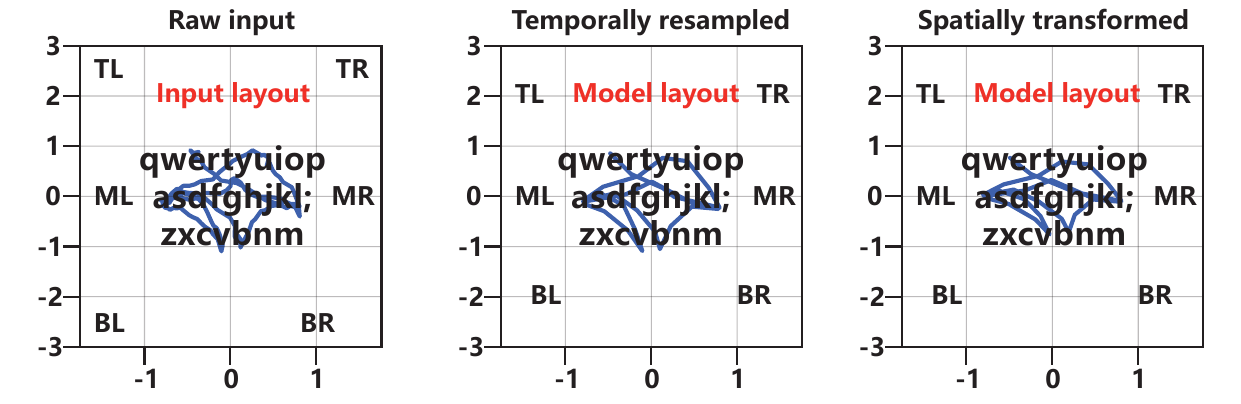}
    \caption{The novel process of converting word-gesture trajectory data from one keyboard layout to another. It demonstrates an example of a trajectory (for the word `available') that undergoes both temporal and spatial transformations. The keyboard layouts vary in terms of key spacing, bottom row shifts, and the presence or absence of the apostrophe key.
    }
    \label{fig:transformation}
\end{figure}

\begin{s_itemize}
    \item \textbf{Training Dataset:} 
    How We Swipe Dataset~\cite{Liu2016How} is a large-scale gesture typing dataset that was collected via a web-based custom virtual keyboard, involving 1,338 users who submitted 11,318 unique English words. However, the dataset used a keyboard layout customized to a mobile phone. 
    Our preliminary analysis revealed that merely normalizing the trajectory data by dividing it by the keyboard's width and height leads to training data that is not viable for effective use.
    Therefore, we employ a novel transformation method which is a piecewise affine transformation to transform the dataset to our keyboard layout. 
    \item \textbf{Piecewise Affine Transformation:}
    This transformation process, illustrated by Figure~\ref{fig:transformation}, adjusts the swipe path onto a standard keyboard layout (known as the model layout) before it is processed by the model. This transformation relies on two corresponding sets of anchor points, one set on the input layout and another on the model layout. In our approach, these anchor points include 1) the central points of all keys that have the same labels across layouts, and 2) an additional six anchor points surrounding the keyboard, each positioned three key distances from the English-letter region's border (denoted by `TL', `ML', `BL', etc. in Figure~\ref{fig:transformation}). These additional points are crucial for adjusting for swipe paths that extend beyond the keyboard border. Once we've established the anchor points, the area they cover is partitioned into a grid. Within each grid subregion, which is enclosed by four nearby anchor points, spatial coordinates are modified in a manner akin to perspective transformation seen in photo editing. 
    This transformation method improves the test accuracy of a word-gesture decoding model, on the word-level gesturing dataset collected from Study 1, by 76\% compared to when using training data processed through the previously mentioned normalization method.
    \item \textbf{Training Model:} 
    We build our gesture decoding model using the Attention-Enhanced Bi-directional LSTM with CTC loss (AE-BLSTM-CTC) architecture proposed by Shen et al.~\cite{shen2023fast}. Then we trained our model on the previously transformed dataset. We used the same training hyperparameters as in Shen et al.~\cite{shen2023fast}. The training hyperparameters for the word-gesture decoding model is directly adopted from Shen et al.~\cite{shen2023fast}. 
\end{s_itemize}

\subsubsection{Spatial Spelling Correction Model}
The word-gesture decoding model is a character-level model that predicts the probability of classes (26 characters plus the blank class) at each timestep of the input trajectory sequence. 
As such, the prediction may contain spelling errors, caused by noise from the model as well as from the user's input, thus necessitating an auto-correction model to correct the misspelled words.
Therefore, we propose a probabilistic edit distance that incorporates keyboard spatial information to address these shortcomings. The computation of this spatial-aware probabilistic edit distance involves three steps:

\begin{s_enumerate}
    \item \textbf{Calculation of the insertion probability $P_{insert}(i)$}. This probability measures the likelihood of an insertion at the $i$-th position of the input string. If the insertion occurs at the end of the input, the probability is $\log(1)$, otherwise, it is equivalent to the omission probability $P_{omit}$. $P_{omit}$ represents the probability that a user omits a character when typing. This is modeled as a logarithmic probability with a base value of 0.06, yielding a logarithmic probability of -1.22.
    \item \textbf{Calculation of the deletion probability $P_{delete}(i)$}. This probability measures the likelihood of a deletion at the $i$-th position of the intent string and is equivalent to the stray probability $P_{stray}$. $P_{stray}$ represents the probability that a user accidentally adds an extra character. This is also modeled as a logarithmic probability with a base value of 0.06, yielding a logarithmic probability of -1.22.
    \item \textbf{Calculation of the substitution probability $P_{sub}(i, j)$}. This probability measures the likelihood of a substitution at the $i$-th position of the intent string and the $j$-th position of the input string. 
    If the $i$-th character of the intent string is equal to the $j$-th character of the input string, the substitution probability is $\log(1)$. Otherwise, the substitution probability is equivalent to the substitution probability $P_{sub}$. $P_{sub}$ represents the probability that a user substitutes one character for another. 
    This measure differentiates between adjacent keys and non-adjacent keys on the keyboard. 
    For adjacent keys (e.g., `q' and `w'), the base probability is 0.17, yielding a logarithmic probability of -0.77. 
    For non-adjacent keys, the base probability is 0.01, yielding a logarithmic probability of -2. 
\end{s_enumerate}

We obtain the base probability through an estimation of character error rates on publicly available experiment data from a mid-air gesture-typing keyboard~\cite{shen2023fast}. The spatial-aware probabilistic edit distance is then calculated as a composite function of these probabilities:

\begin{align*}
P_{ED} &= P_{omit}^{n_{ins}} \cdot P_{stray}^{n_{del}} \cdot P_{sub}(s_1) \cdot P_{sub}(s_2) \\
\intertext{Taking the logarithm of both sides, we get:}
\log(P_{ED}) &= n_{ins} \cdot \log(P_{omit}) + n_{del} \cdot \log(P_{stray}) \\
&\quad + \log(P_{sub}(s_1)) + \log(P_{sub}(s_2))
\end{align*}

where $n_{ins}$ and $n_{del}$ denote the number of insertions and deletions, respectively, and $s_1$ and $s_2$ denote the substitutions.

\subsubsection{Contextual Language Model}
\begin{figure}[t]
    \centering
    \includegraphics[width=0.85\textwidth]{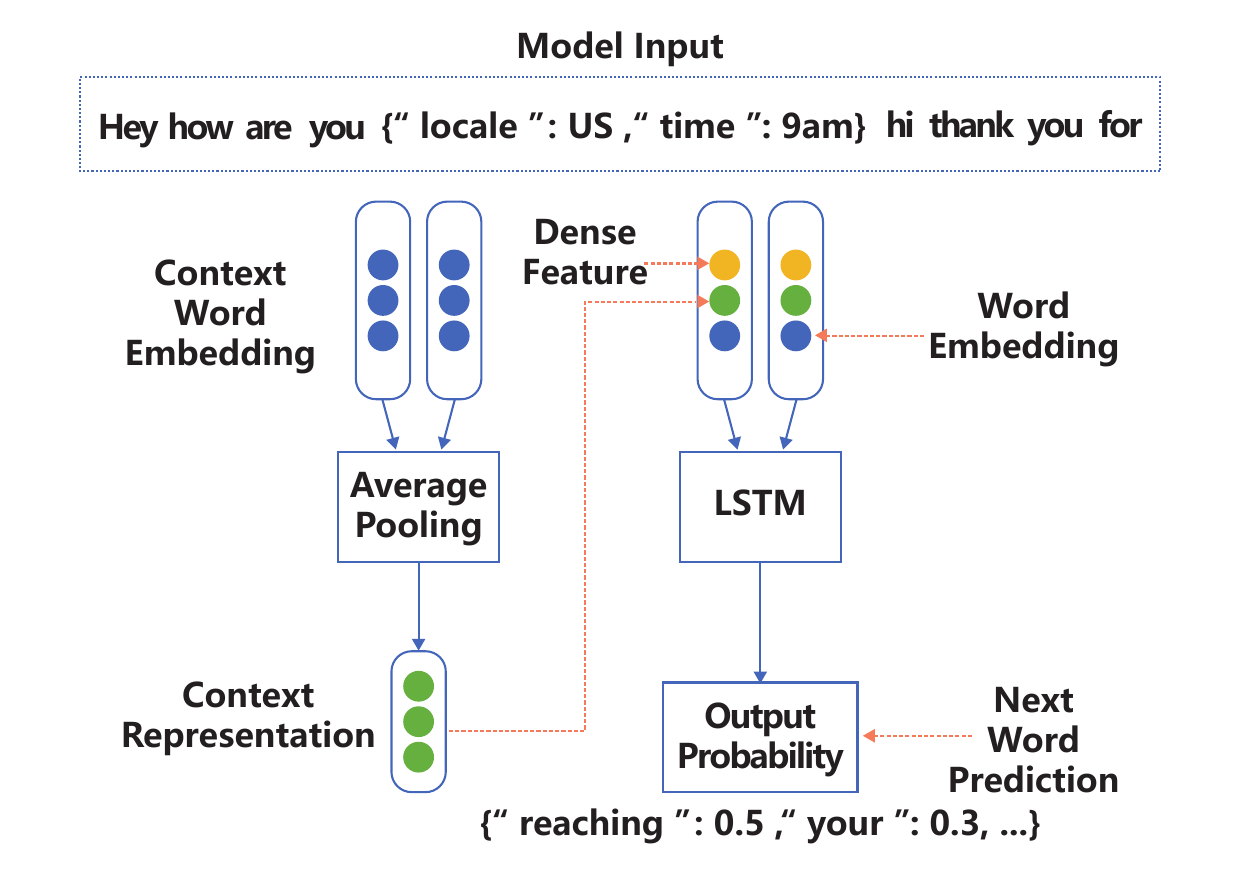}
    \caption{Contextual LSTM-based language model structure, with model input including a) previous conversation history (eg. `Hey how are you'), b) context tags (eg. `location': US, `time':9am), c) previously entered text (eg. `hi thank you').
    This model predicts the next word based on probabilities. It assigns a probability to each word in the corpus; for example, the probability of `reaching' is 0.5, `your' is 0.3, and there are smaller probabilities for many other words. }
    \label{fig:context_lm}
\end{figure}

\begin{table}[t]
\scriptsize
\caption{Experiments for contextual language model. Perplexity is a measure of how well a probability model predicts a sample, with lower values indicating better predictive accuracy.}
\centering
\begin{tabular}{lc}
\hline
\textbf{Language Model Architecture} & \textbf{Perplexity} \\
\hline
Baseline Bi-LSTM LM & 70.91 \\
Contextual Bi-LSTM LM & 42.82 \\
Pre-trained Contextual Bi-LSTM LM & 37.16 \\
\hline
\end{tabular}
\label{tab:model_perplexity}
\end{table}

We employed a bi-directional LSTM (Bi-LSTM)~\cite{SchusterPaliwal1997,Salehinejad2017,Graves2005} model to generate predictions of subsequent word based on previous tokens. 
The input and output of the model are illustrated in Figure~\ref{fig:context_lm}.
Our model structure comprises four key components: an embedding layer, a representation layer, a decoder layer, and a contextual encoder.
\begin{s_enumerate}
    \item Embedding Layer: We utilize a Convolutional Neural Network (CNN)~\cite{Venkatesan2017} for our embedding layer with an embedding dimension of 38, 100 kernels of size 3. 
    We opted against using dilation and weight normalization~\cite{Salimans2016WeightNA} in the CNN to retain the original characteristics of the input data. These hyperparameters were determined through a grid-search-based hyperparameter optimization process.
    \item Representation Layer: For this layer, we employ a BiLSTM. 
    % This layer is designed to understand the context and dependencies in the input data.
    Our BiLSTM has two layers and a dimension of 2048.
    We incorporate a dropout of 0.001 to prevent overfitting.
    \item Decoder Layer: The final layer of our model is an MLP decoder.
    This layer transforms the high-level features learned by the previous layers into the final output. 
    Our MLP has a hidden dimension of 1024 and leverages ReLU as the activation function. A dropout rate of approximately 0.00092 is used to further mitigate overfitting.
    \item Contextual Encoder: To efficiently incorporate long context information, we introduced a contextual encoder. This encoder first performs average pooling on the context word embeddings and then concatenates these with the word embeddings. The contextual encoder is co-trained with the remaining language model (LM) modules. 
\end{s_enumerate}

We employ perplexity, as defined by \cite{Shannon1948}, to assess the performance of language models. This metric is calculated as the exponentiation of the average negative log-likelihood of the test set words, normalized by the number of words.  As evident from Table~\ref{tab:model_perplexity}, integrating contextual information significantly enhances model perplexity. 
Furthermore, pre-trained language models, as highlighted by \cite{devlin2019bert} and \cite{vaswani2017attention}, demonstrate exceptional utility in scenarios where training data is significantly limited. 
Our approach involved initially pre-training the language model using diverse sources such as public comments and posts~\cite{reddit_dataset,yelp_dataset,enron_email_dataset,wikipedia_talk_pages}, followed by fine-tuning on the training dataset outlined by \cite{shen2022kwickchat}. 
As demonstrated in Table~\ref{tab:model_perplexity}, the pre-trained contextual language model substantially outperforms basic models, thereby validating the effectiveness of pre-training coupled with subsequent fine-tuning. 
After quantization, the final exported contextual language model is only 7MB, enabling real-time execution on contemporary mobile phone processors.

\subsubsection{Implementation Details}
We use PyText~\cite{pytext} to implement the contextual language model. We employ the Adam optimizer to train our model, with a learning rate of 0.001, epsilon of 1e-8, and weight decay of 0.00001. The model is trained for a total of 25 epochs, with an early stopping criterion set after 5 epochs.
% These hyperparametrs was found through hyperparmaeters sweeiping .
We accumulate gradients over 4 batches before updating the model parameters, and each epoch consists of 4700 such batches. Our model is designed to leverage distributed training with a world size of 8, effectively utilizing multiple GPUs to speed up the training process. 
The \textit{Score Fusion} framework combines the log probabilities from the three models to assign a score to each word, creating a list of suggestions. More specific details are illustrated by Algorithm~\ref{alg:score_fusion}.

\begin{algorithm}[t]
\scriptsize
\caption{\textit{Score Fusion}}
\begin{algorithmic}[1]
\Require Trajectory, SwipeCorrectionCoeff, LmCoeff, NumSuggestions, vocab, context
\Ensure Sorted suggestions
\State $raw\_decodings \gets WordGesture\_Decoder(Trajectory)$
\State Initialize $suggestions$ as an empty dictionary
\For{each $raw\_word$, $raw\_score$ in $raw\_decodings$}
    \State $text\_probabilities \gets Context\_Language\_Model(context)$
    \State $typo\_probabilities \gets Spatial\_Spelling\_Correction(raw\_word)$
    \For{$i$ in $0$ to $len(typo\_probabilities) - 1$}
        \State $correction \gets vocab[i]$
        \State $index \gets find(correction, vocab)$
        \State $lm\_score \gets text\_probabilities[index]$
        \State $blended\_score \gets (1 - SwipeCorrectionCoeff - LmCoeff) * raw\_score + LmCoeff * lm\_score + SwipeCorrectionCoeff * typo\_probabilities[i]$
        \If{$correction$ in $suggestions$}
            \State $blended\_score \gets max(suggestions[correction], blended\_score)$
        \EndIf
        \State $suggestions[correction] \gets blended\_score$
    \EndFor
\EndFor
\State $sorted\_suggestions \gets sort(suggestions, byValue, descending)$
\If{$NumSuggestions < len(sorted\_suggestions)$}
    \State $sorted\_suggestions \gets sorted\_suggestions[0:NumSuggestions]$
\EndIf
\Return $sorted\_suggestions$
\end{algorithmic}
% \caption{Algorithm for \textit{Score Fusion}.}
\label{alg:score_fusion}
\end{algorithm}

\begin{figure}[t]
    \centering
    \includegraphics[width=0.5\textwidth]{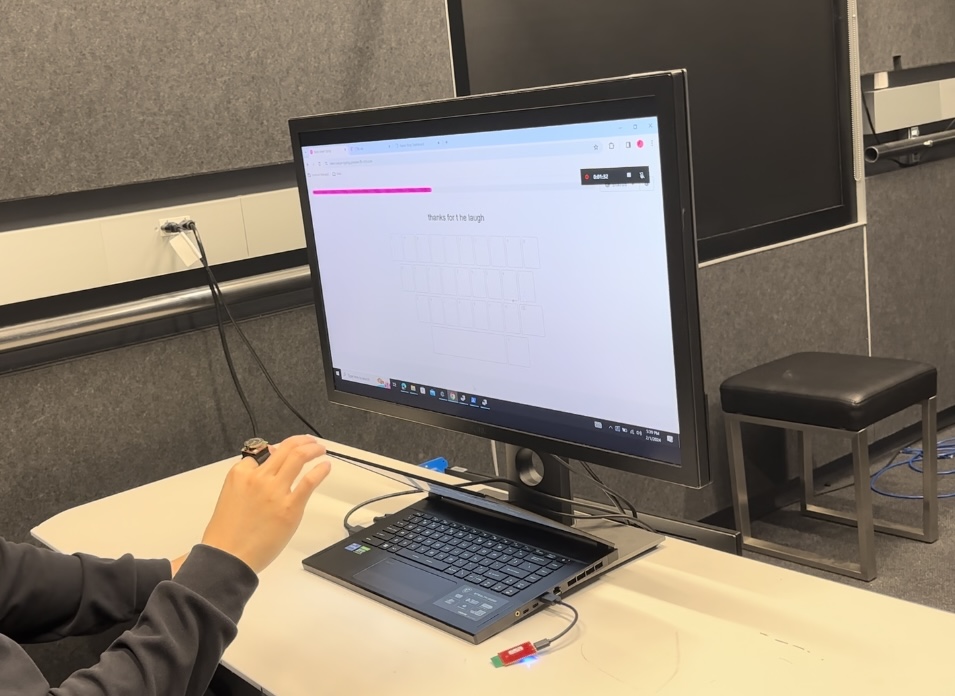}
    \caption{Experiment Setup for Study 2: A participant is seated in front of a monitor, wearing a ring on their index finger. They can comfortably rest their arms on the armrests of the chair and freely move their wrist to perform mid-air gesture typing.}
    \label{fig:study_setup}
\end{figure}

\section{User Study 2: Longitudinal Evaluation of RingGesture}
Our second study was driven by two primary objectives. 
First, we compared two text entry conditions: one that used \textit{Score Fusion} when entering phrases and another that used \textit{Naive Correction}.  
Second, we evaluated the potential text-entry performance of \textit{RingGesture}, which involved assessing both the users' initial proficiency and their progress over time. 
Through the analysis of the learning curve, we aimed to gain a deeper understanding of the system's usability and the time required for users to reach proficiency.

\begin{s_enumerate}
    \item \textbf{Participants}: 
    We recruited 16 volunteers as participants through an internal mailing list. 
    The details of their demographics are listed below. 
    13 participants are right-handed, and 3 participants are left-handed.
    The ages of the participants ranged from 21 to 49.5 years, with an average age of 37.72 years. The standard deviation in the age distribution was 10.30 years. In terms of gender, 6 participants (37.5\%) identified as male, 9 participants (56.25\%) identified as female, and 1 participant (6.25\%) preferred not to disclose their gender. Among the participants, 12 (75\%) reported they would be wearing the device on their right arm, while the remaining 4 (25\%) would wear it on their left arm. 
    When asked about the frequency of using gesture typing, 5 participants (31.25\%) reported always using it (at least once a day), 7 participants (43.75\%) sometimes (at least once a week), 2 participants (25\%) seldom (less than once a month), and the remaining 2 participants never used it. 
    \item \textbf{Phrase Set}:
    The studies utilize the phrase set derived from the ConvAI2 challenge dataset~\cite{burtsev2018first}, which consists of a total of 42,612 unique phrases. Each unique phrase is accompanied by two additional elements: context tags, which include speaker persona, and conversation history.
    For example, one unique phrase in the dataset might be `I read books in the afternoon.' This phrase would be accompanied by context tags such as `love reading' for speaker persona, and a conversation history element like `How are you?'. 
    The dataset's unique characteristics, namely its conversational basis and inclusion of persona information, make it an ideal tool for evaluating the contextual capabilities of intelligent text entry systems.
    In the study, the contextual information, which is pre-defined with the stimulus phrase, is fed automatically as additional input to the word error correction frameworks. 
    Allowing participants to freely enter text and use their own conversational language necessitates a large-scale, in-the-wild study to ensure a fair comparison between the two conditions. However, our current implementation of the ring device does not support running such a large-scale study in a natural setting for the comparison of these two conditions.
    By pre-establishing the conversational context, our study offers valuable insights into the realistic text entry rates of upcoming systems that will account for the historical context of use.

    \item \textbf{Apparatus}:
    Participants controlled the cursor using the ring, and the cursor was displayed on a virtual keyboard on a monitor. The delimitation is performed when detecting a pinch. The monitor was connected to a Lenovo PC (ThinkStation) equipped with an Intel Xeon processor. 
    We chose a computer over AR glasses to allow demonstrators and participants to view the same screen. 
    This setup enabled participants to pose questions and receive immediate feedback about the on-screen scenes, and provided a more effective platform for demonstrators to explain the swiping and delimitation techniques during the practice stage.

    \item \changetext{\textbf{Baseline}:
    \textit{Naive Correction} acts as the baseline in this study. 
    It is a state-of-the-art word prediction framework for gesture typing that was used in Shen et al.~\cite{shen2023fast}, as illustrated by Figure~\ref{fig:sequential_correction}. }
    
    \item \textbf{Procedure}:
    Each participant was seated before a computer screen displaying a keyboard interface as illustrated in Figure~\ref{fig:study_setup}. Participants initially practiced with a set of 5 phrases, during which they were encouraged to ask any questions. 
    Subsequently, they completed four sessions, each consisting of four blocks. 
    Each block contained two conditions: with \textit{Score Fusion} and with \textit{Naive Correction}.
    In each condition, participants input 10 phrases, resulting in a total of 20 phrases per block. 
    The conditions were counterbalanced, and the sessions were scheduled across two weeks on separate days.
    In each of the four sessions, participants were given time to familiarize themselves with the functionality of \textit{RingGesture} before beginning their typing. 
    % The procedure of the study is illustrated by Figure~\ref{fig:procedure}.
    % On average, participants typed 1.2 phrases at the beginning of each session. 
    During each session, participants were instructed to type the phrases `as swiftly and accurately as possible, as if typing an email to a colleague.' A break of up to three minutes was allowed between each block of 20 phrases. Similar to Study 1, we also use \texttt{Space Bar} to proceed to the next phrase. The average duration of each session was approximately 30 minutes.
    At the end of the study, participants were invited to complete a post-study questionnaire similar to Study 1. This included a Likert scale rating on the same six aspects in study 1: \textit{Ease of Typing/Effort}, \textit{Ease of Learning}, \textit{Perceived Speed}, \textit{Perceived Accuracy}, \textit{Hand Fatigue} and \textit{Eye Fatigue/Attention Switch}.
    Additionally, participants were requested to complete a standardized System Usability Score form~\cite{Brooke1996}. Finally, they were posed the open-ended question: `Did wearing the ring influence your swipe behavior or task performance?'
\end{s_enumerate}

\subsection{Performance Analysis}

\begin{figure}[t]
    \centering
    \begin{subfigure}{0.45\textwidth}
        \centering
        \includegraphics[width=0.98\linewidth]{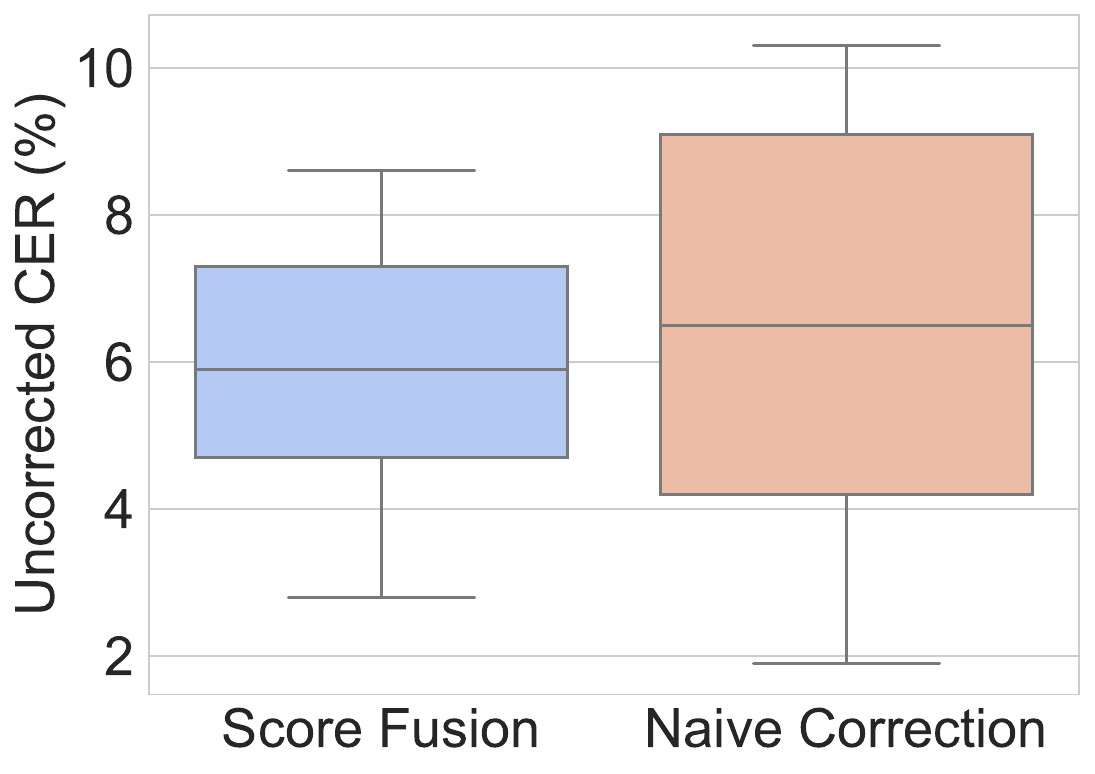}
        \caption{Uncorrected Character Error Rate}
        \label{fig:study2_ucer}
    \end{subfigure}%
    \hfill
    \begin{subfigure}{0.45\textwidth}
        \centering
        \includegraphics[width=0.98\linewidth]{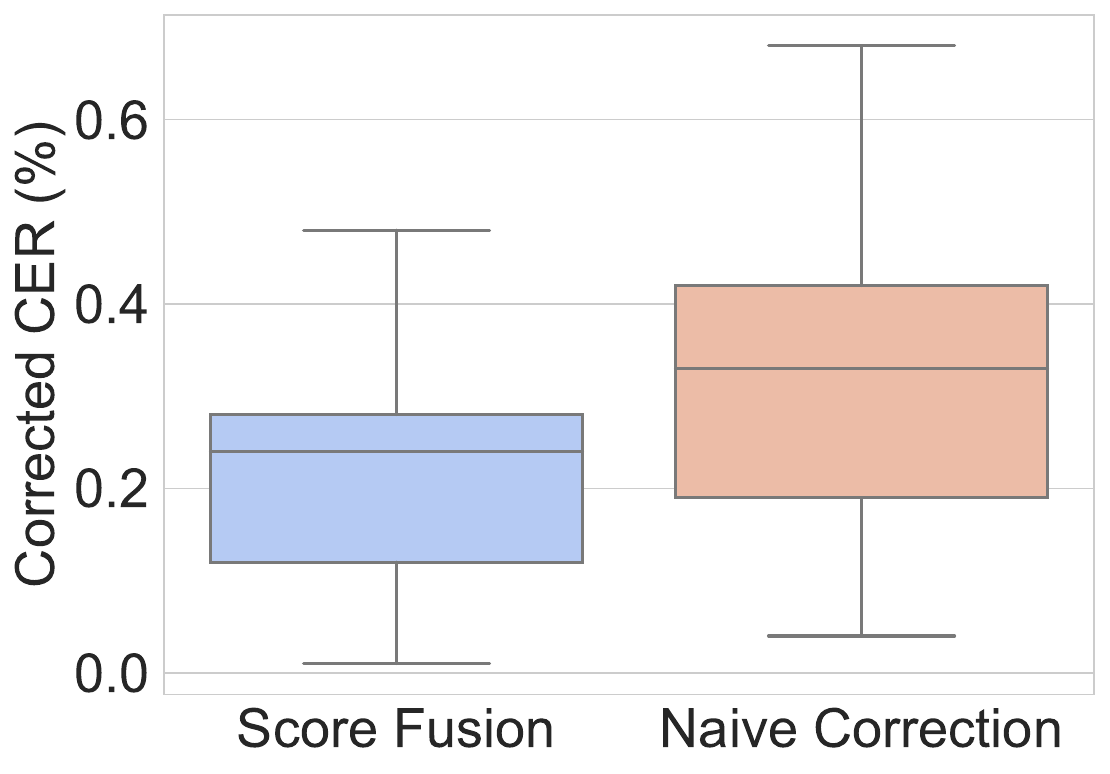}
        \caption{Corrected Character Error Rate}
        \label{fig:study2_ccer}
    \end{subfigure}
    \caption{Box plots depicting mean, median and quartiles of the participants’ performance including corrected and uncorrected character error rates under the two conditions.}
    \label{fig:study2_error_rates}
\end{figure}

\begin{figure}[t]
    \centering
    \includegraphics[width=0.7\textwidth]{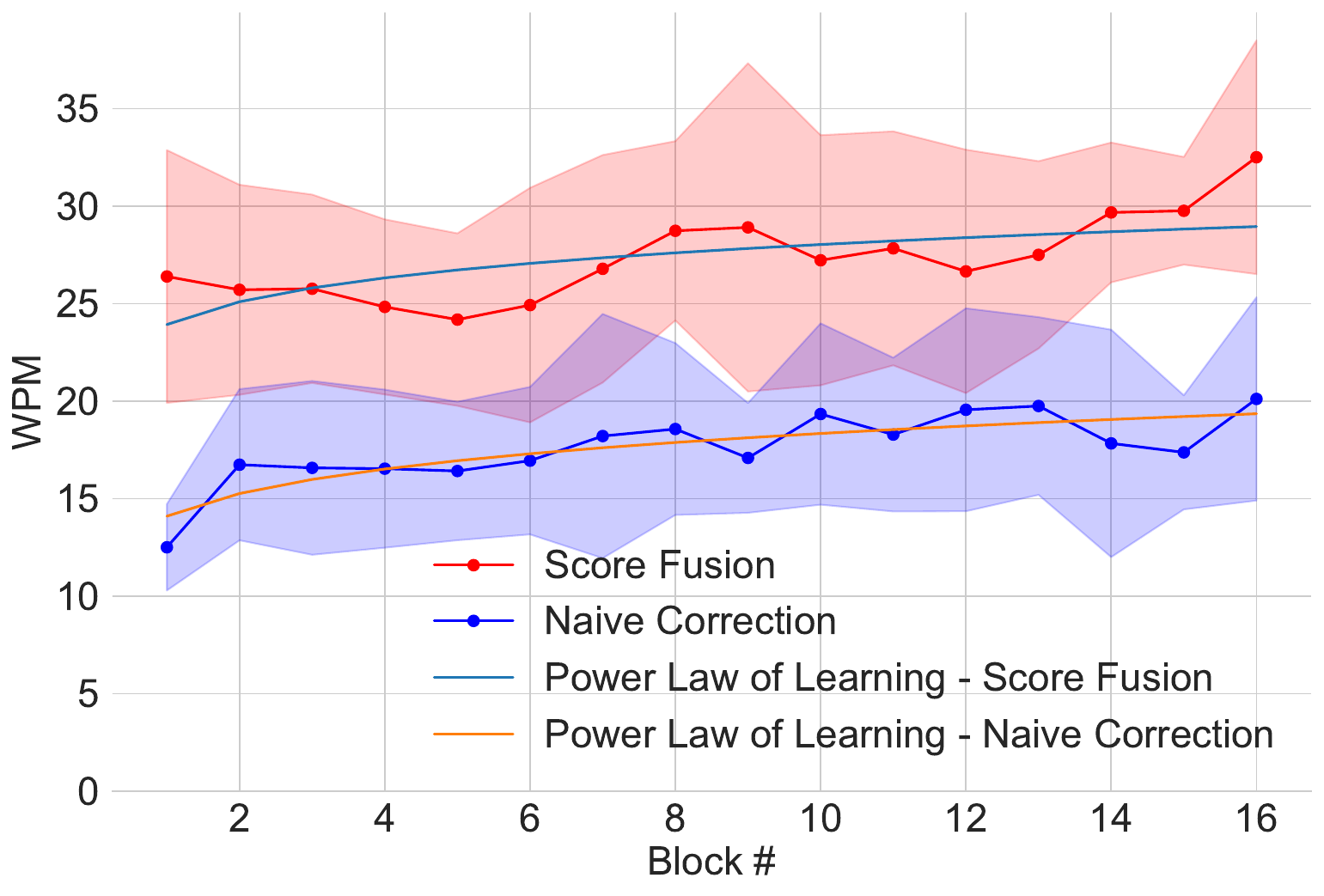}
    \caption{Comparative analysis of \textit{Score Fusion} and \textit{Naive Correction} conditions over 16 longitudinal blocks.}
    \label{fig:study2_result}
\end{figure}

\begin{figure}[t]
    \centering
    \begin{subfigure}{0.45\textwidth}
        \centering
        \includegraphics[width=0.98\linewidth]{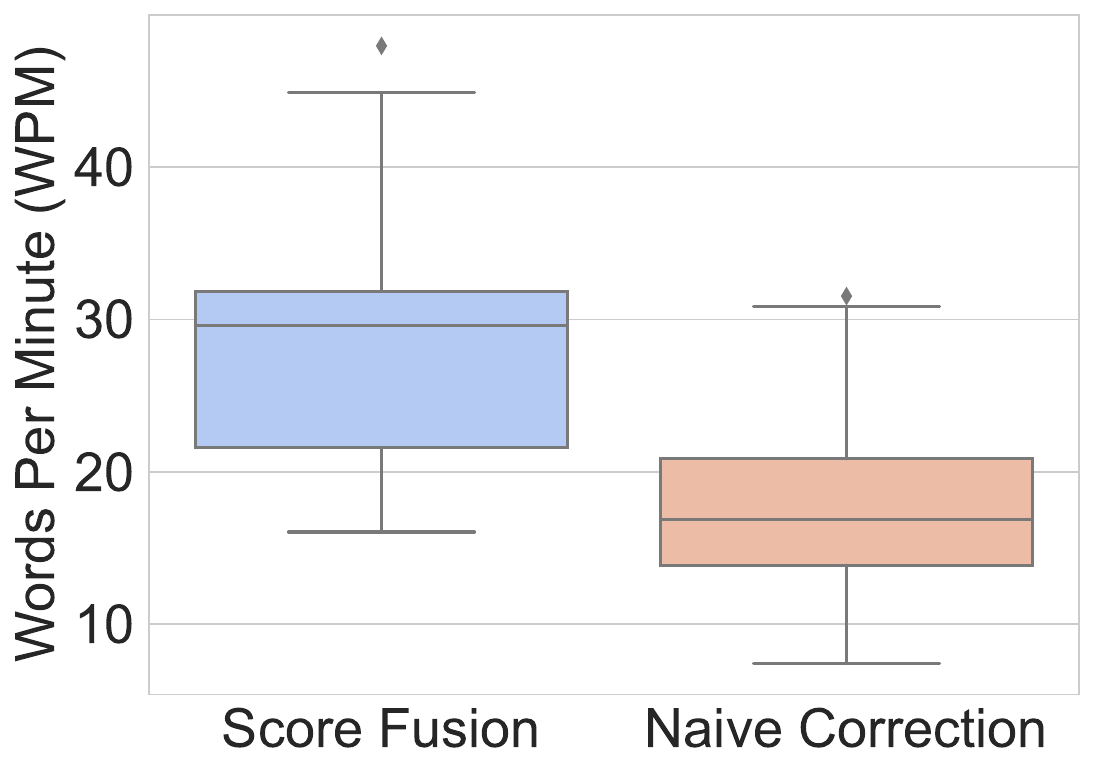}
        \caption{Comparison between the two conditions: \textit{Score Fusion} and \textit{Naive Correction}.}
        \label{fig:study2_compare}
    \end{subfigure}%
    \hfill
    \begin{subfigure}{0.45\textwidth}
        \centering
        \includegraphics[width=0.98\linewidth]{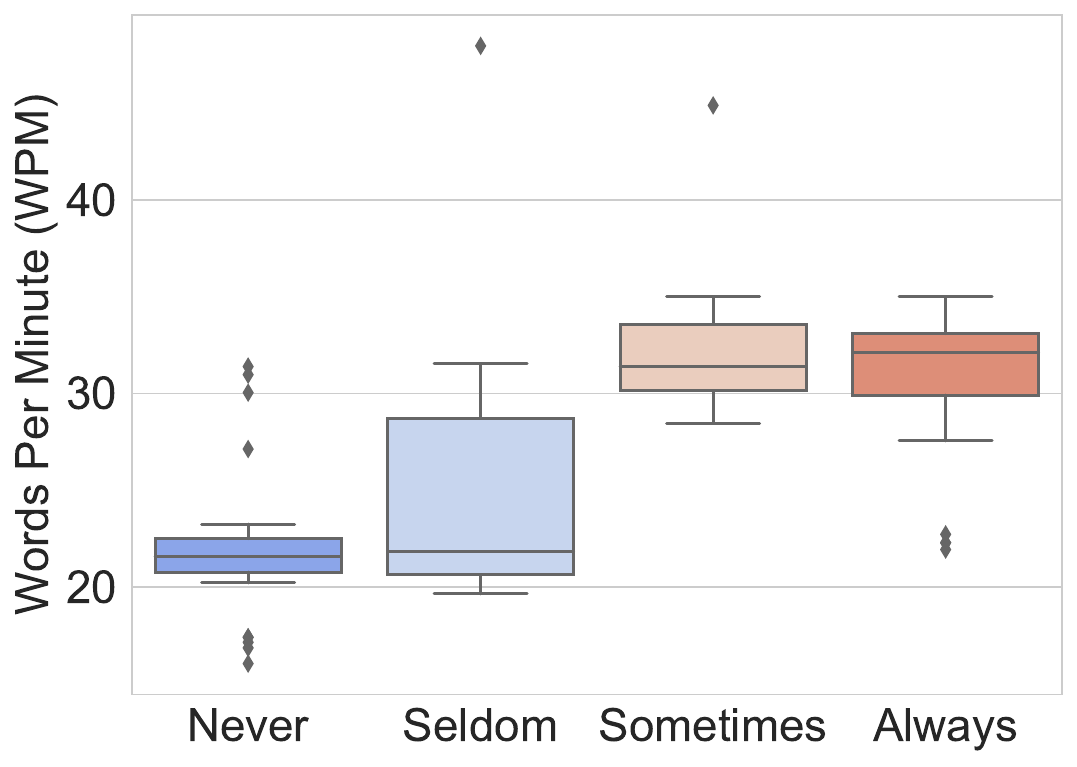}
        \caption{Performance of participants with different gesture typing proficiency under \textit{Score Fusion} condition.}
        \label{fig:study2_frequency_speed}
    \end{subfigure}
    \caption{Box plots depicting mean, median, and quartiles of the text entry rate performance in Study 2 showing overall entry rates (a) and by self-assessed experience level (b).}
    \label{fig:study2_entry_rates}
\end{figure}

\begin{table}[t]
\scriptsize
\centering
\caption{ANOVA-RM results for assessing the impact of condition and block on the text entry rate across repeated measures.}
\begin{tabular}{lcccc}
\hline
Factor          & F Value & Num DF & Den DF & Pr $>$ F  \\
\hline
Condition       & 30.276 & 1.000 & 7.000 & 0.001  \\
Block           & 4.108  & 15.000 & 105.000 & 0.000  \\
Condition:Block & 1.333  & 15.000 & 105.000 & 0.196  \\
\hline
\end{tabular}
\end{table}

\subsubsection{Error Rate Analysis}
Figure~\ref{fig:study2_ucer} and Figure~\ref{fig:study2_ccer} shows that \textit{Score Fusion} also has a lower and more consistent uncorrected and corrected CER compared to \textit{Naive Correction}, indicating it's a more effective method for error prevention before correction is even applied and thereby reducing errors. 
Overall, \textit{Score Fusion} outperforms \textit{Naive Correction} by 28.2\% in uncorrected CER, and 29.0\% in corrected CER.
% \item \textbf{Entry Rate Analysis:}

\subsubsection{Entry Rate Analysis}
Figure~\ref{fig:study2_result} illustrates the performance of the \textit{RingGesture} system when coupled with the \textit{Score Fusion} framework. The system achieved an average entry rate of 27.3 words per minute (WPM), starting at 26.4 WPM in the initial block and rising to 32.5 WPM in the final block. This progression showcases the improvement from novice to expert levels of performance.

Figure~\ref{fig:study2_compare} indicates that \textit{Score Fusion} shows a higher median entry rate and a tighter interquartile range compared to \textit{Naive Correction}, suggesting it enables faster text entry and provides more consistent performance across different blocks or users.
Figure~\ref{fig:study2_frequency_speed} demonstrates a general trend of increasing entry rates with more frequent use. 
The spread of entry rates (as shown by the interquartile ranges and outliers) also seems to generally decrease with more experience, indicating that users become not only faster but also more consistent with practice.

Additionally, we use Repeated Measures ANOVA (RM-ANOVA)~\cite{Maxwell2004DesigningEA} and the power law of learning~\cite{Snoddy1926} to analyze gesture typing performance under two conditions: \textit{Score Fusion} and \textit{Naive Correction}. RM-ANOVA is chosen for its efficacy in handling within-subject variance across repeated observations, allowing us to assess the impact of the two conditions over time and the consistency of participant performance. We also use the power law of learning to get insight into improvement rates and learning dynamics, improving our understanding of how participants adapt to each condition. 

\begin{s_itemize}
    % \item \textbf{Error Rate Analysis:}
    \item \textit{Between Condition}:
    The RM-ANOVA analysis revealed a significant main effect of condition (F=30.276, p=0.001), indicating a substantial difference in typing performance between the \textit{Score Fusion} and \textit{Naive Correction} conditions. This statistical significance is further indicated by the performance metrics, where \textit{Score Fusion} exhibited a mean text entry rate of 27.3 WPM, outperforming \textit{Naive Correction}'s mean of 17.6 WPM. This difference translates to a notable 55.2\% improvement in favor of\textit{Score Fusion}, emphasizing not just a statistical but a practical superiority in typing efficiency. 
    \item \textit{Between Blocks}:
    Additionally, the RM-ANOVA showed a significant effect for blocks (F=4.108, p=0.000), suggesting variability in typing performance over time which could be caused by learning effects, yet no significant interaction between condition and block (F=1.333, p=0.196) was observed, indicating that the performance advantage of \textit{Score Fusion} is consistent across different time points. This consistency, backed by \textit{Score Fusion}'s superior statistics (with a standard deviation of 5.9, minimum of 16.1, and maximum of 47.9) compared to \textit{Naive Correction}'s (standard deviation of 4.7, minimum of 7.4, and maximum of 31.5), highlights how different word prediction frameworks in gesture typing not only influence overall performance but also ensure sustained efficiency across blocks.
    \item \textit{Power Law of Learning}:
    As there is a significant variability in typing performance over time, we analyze the learning dynamics through the power law of learning for \textit{Score Fusion} and \textit{Naive Correction} conditions. Figure~\ref{fig:study2_result} also plots the power law of learning for the two conditions.
    We find distinct patterns in participants' improvement rates. The R-squared values are 0.415 for \textit{Score Fusion}, and 0.697 for \textit{Naive Correction}, suggesting that learning under \textit{Naive Correction} is slightly more predictable over time than under \textit{Score Fusion}. 
    Initial performance levels, indicated by \(a = 23.9\) for \textit{Score Fusion} and \(a = 14.1\) for \textit{Naive Correction}, show that participants start off better with \textit{Score Fusion}. 
    However, the rate of learning, represented by \(b\), is faster in \textit{Naive Correction} (\(b = 0.115\)) than in \textit{Score Fusion} (\(b = 0.069\)), despite the higher initial performance in the latter. 
    The more predictable learning effect, and the higher learning rate might be caused by participants' adaptation to the \textit{Naive Correction} framework, whereas \textit{Score Fusion} offers accurate predictions, eliminating the need for user adaptation.
    % This nuanced insight, coupled with the precision of parameter estimates as indicated by the covariance matrices, highlights a subtle balance between initial efficiency and learning progression in gesture typing methodologies.
\end{s_itemize}

\subsection{Subjective Ratings \& Feedback}
% \label{sec:study2_aspects}

\begin{figure}[t]
    \centering
    \includegraphics[width=0.7\textwidth]{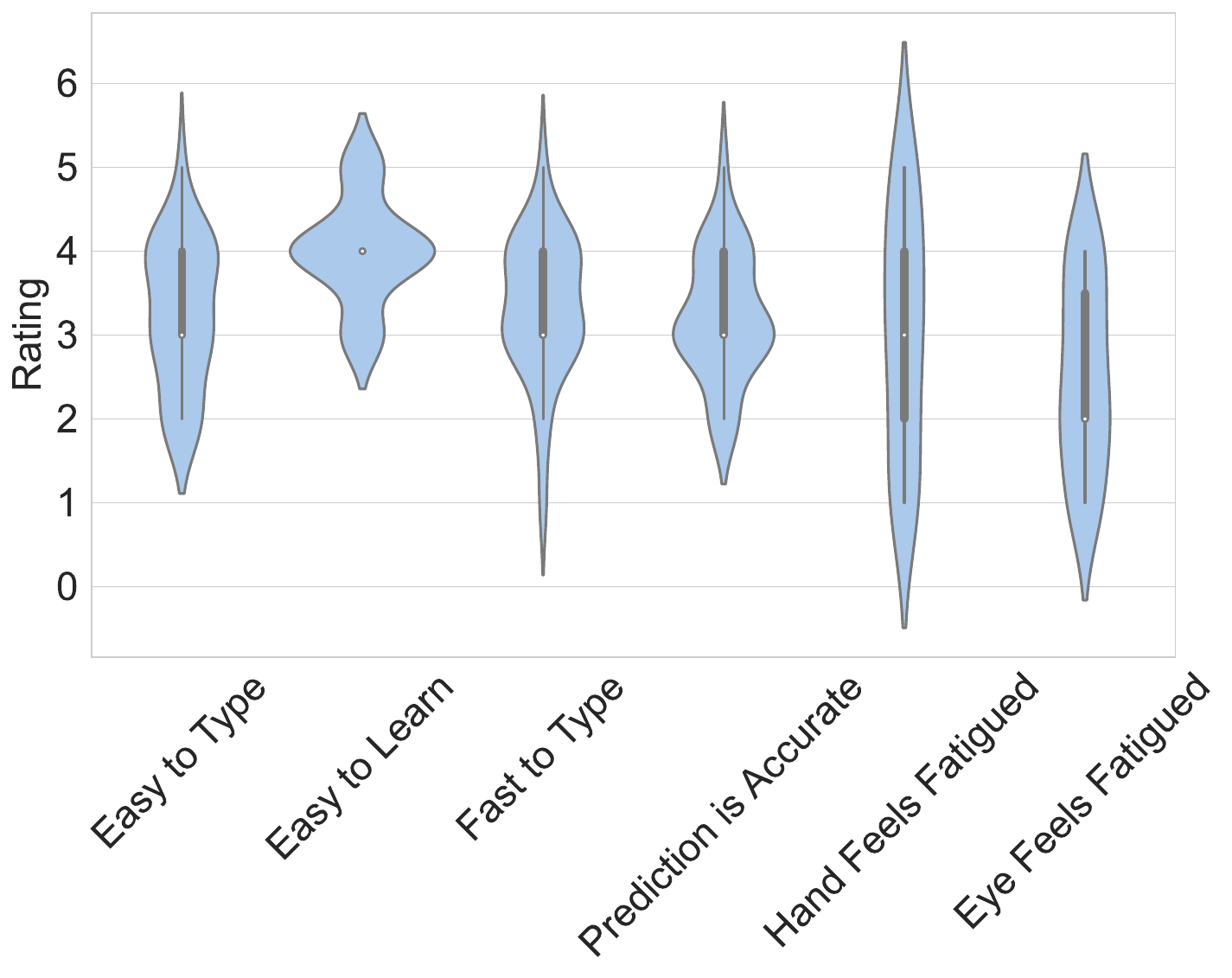}
    \caption{    Violin plots of answers to subjective rating questions scored on 5-point Likert scales. 
    % Violin plots are modified box plots that add estimated kernel density plots to the summary statistics displayed by box plots.
    The 5-point Likert scales ranged from 1 (strongly disagree) to 5 (strongly agree).
    }
    \label{fig:study2_response}
\end{figure}
% \subsubsection{Ratings of Six Aspects}

\begin{s_itemize}
\item \textbf{System Usability Score}: 
The overall System Usability Scale (SUS) score for the gesture typing system is 83. This indicates that users found the system to be highly usable.
% The SUS score, a reliable tool for measuring the usability of various systems, typically ranges from 0 to 100. 
A score above 68 is considered above average, and anything above 80 is an indication of excellent usability. In this context, a score of 83 suggests that the gesture typing system was not only easy to use but also met or exceeded the expectations of most users in terms of efficiency and satisfaction.
This score aligns with the ratings of the six aspects, suggesting that the gesture typing method is easy to learn and efficient to use.
\item \textbf{Ratings of Six Aspects}: 
The post-study questionnaire used a Likert scale to assess six aspects, as illustrated in Figure~\ref{fig:study2_response}. Participants generally found gesture typing moderately easy, with an interquartile range (IQR) from 3 to 4, suggesting a moderate consensus among them.
The IQR for ease of learning is narrow, centered on a score of 4, implying minimal variance and indicating that participants found the method easy to learn.
The distribution of ratings on perceived speed is broader, indicating greater variability in how participants perceive the speed of gesture typing. The majority of users perceived gesture typing as a relatively quick input method, which aligns with our previously discussed quantitative results. 
Perceived accuracy has a distribution similar to that of perceived speed, with few participants rating as having low accuracy. This suggests a contribution from the intelligent text entry framework, \textit{Score Fusion}, in providing accurate text entry predictions.
Additionally, many reported minimal hand fatigue, suggesting that the ring-based gesture typing may offer an ergonomic advantage.
The plot also indicates that participants had neutral feelings regarding eye fatigue. 
\item \textbf{Feedback for Improvement}:
While we received many positive comments and overall feedback sentiment, we also recognize areas for improvement. Several participants highlighted the need for refinement. Specifically, \textit{Participant 3} shared, `I had to concentrate to keep my middle finger out of the way,' due to the system's method of detecting a pinch gesture, which essentially requires connecting the thumb and index finger to form a closed loop. If the middle finger inadvertently touches the index finger and the sensor signal change surpasses the threshold, the system mistakenly registers this as a pinch gesture, leading to inaccuracies. However, such incidents were infrequent, observed only on rare occasions with \textit{Participant 3} and \textit{Participant 8}.

\end{s_itemize}

\subsection{Ablation Analysis of Score Fusion Components}
To understand the contribution from each component in the \textit{Score Fusion} framework, we performed an ablation analysis on the logged word-trajectory from the Study 2.
% with the aim of investigating the contribution from each of the components in the \textit{Score Fusion} algorithm.
The results are summarized in Table~\ref{table:ablation_study}.
These results underscore the significance of each component, as each addition led to substantial improvements in accuracy. 

\begin{table}[t]
\scriptsize
\caption{Ablation analysis of \textit{Score Fusion}'s components using the dataset logged from Study 2, with outcomes reported as Character Error Rate (CER). Components include GDM (Gesture Decoding Model), SSCM (Spatial Spelling Correction Model), and CLM (Contextual Language Model).}
\centering
\begin{tabularx}{0.9\linewidth}{Xl}
\hline
\textbf{Components} & \textbf{Character Error Rate} \\
\hline
GDM & 27.86\% $\pm$ 4.35\% \\
GDM + SSCM & 12.12\% $\pm$ 3.67\%\\
\textbf{GDM + SSCM + CLM} & \textbf{5.56}\% $\pm$ 1.33\%\\
\hline
\end{tabularx}
\label{table:ablation_study}
\end{table}

\section{Discussion}
In this paper, we have presented \textit{RingGesture}, a novel ring-based text entry system for lightweight AR glasses. 
\textit{RingGesture} incorporates an intuitive ring-based mid-air pointing and selection technique that allows users to perform mid-air gesture typing. 
It also introduces a deep learning word prediction framework, \textit{Score Fusion}, that significantly enhances text entry accuracy and speed.
Through two studies, we have demonstrated the effectiveness and usability of \textit{RingGesture}. 
Study 1 revealed that word-level gesture typing was preferred by users over phrase-level gesture typing. 
\changetext{
Study 2 demonstrated that \textit{RingGesture}, particularly when utilized with the \textit{Score Fusion} framework, facilitates efficient one-handed text entry, achieving text entry rate of 27.3 WPM. This performance is comparable to mobile phone gesture typing, which also averages around 30 WPM~\cite{reyal2015performance}, despite the challenges posed by noise and drift in IMU tracking.
% Study 2 showed that \textit{RingGesture}, especially with the  \textit{Score Fusion} framework, enables efficient one-handed text entry with entry rates approaching 30 WPM, despite the noise and drift associated with IMU tracking, \textit{RingGesture} achieves a text entry rate of 27.3 WPM.
% This rate is similar to mobile phone gesture typing, which is around 30 WPM~\cite{reyal2015performance}.
% It is important to note that our work does not directly replicate the experimental settings of previous methods such as Vulture. Instead, these comparisons serve as reference points. 
When using the conventional word prediction framework \textit{Naive Correction}, the average text entry rate for \textit{RingGesture} drops to 17.6 WPM. 
% The decrease in performance with \textit{RingGesture} and \textit{Naive Correction} can be attributed to the less precise tracking from the ring's IMU compared to OptiTrack. 
The enhanced performance with \textit{RingGesture} and \textit{Score Fusion} is due to \textit{Score Fusion}'s accurate word prediction ability, which effectively compensates for the tracking limitations and maintains high performance. 
The longitudinal evaluation also indicated that users can quickly learn and improve their proficiency with the system over time.
% These results are contextualized within the specific conditions of our study, reflecting the differing environments of prior works.
}

\section{Limitations and Future Work}
% The current ring is still not in its minimal size, which could impact the performance and comfort. 

\changetext{
Our system is specifically designed for text entry on lightweight AR glasses equipped with MicroLED technology. A lightweight AR glass creates a virtual screen at a specific distance in front of the user, offering only 3DOF experience. Viewing a traditional monitor, which is placed directly in front of the user and has a fixed screen position, closely mimics the experience of wearing these AR glasses. Additionally, the current iteration of lightweight AR glasses faces battery life limitations, posing challenges for conducting extended user studies. This similarity in the viewing experience supports the argument that using a monitor as a proxy in user studies can effectively replicate the visual setup of these AR glasses. The use of monitors for conducting text entry studies on time-machine heads-up displays has been implemented in several studies~\cite{gu2020qwertyring, Vulture2014Markussen}. However, we acknowledge that this assumption holds true primarily in controlled lab settings and may not extend to real-world scenarios, especially when the user is in motion. Therefore, we plan to assess the \textit{RingGesture} system in real-life, once lightweight AR glasses are enhanced with longer battery life and become more readily usable, as part of our follow-up work. This will involve assessing the performance of the \textit{RingGesture} system in actual AR experiences, particularly in ``in the wild'' settings such as typing while walking, in a car, or lying on a bed.
}

% Despite the promising results, one limitation is that our study was conducted in a controlled lab setting, which may not fully reflect real-world usage scenarios. 
% Future work should explore the performance and user experience of \textit{RingGesture} in more diverse and realistic environments, such as in the presence of distractions or while the user is in motion. 
The \textit{Score Fusion} framework has the potential to be applied to other decoding methods beyond gesture typing, such as touch typing decoding. 
Its ability to integrate multiple probabilistic models to enhance word prediction accuracy could benefit various text entry systems. 
While we did not test \textit{Score Fusion} with other text entry techniques, we consider this an avenue for future work.
% Lastly, \textit{RingGesture} could also be well-suited for touch typing. 
% In this paper, we did not conduct a comparative longitudinal study to compare these two methods within a single system. As for mobile phones, both text entry methods present their own uniqueness; while gesture typing may offer faster speeds while in motion, touch typing provides better accuracy for out-of-vocabulary words~\cite{reyal2015performance}. 
% This ring-based touch typing technique requires further investigation.

\section{Conclusion}

\textit{RingGesture} presents a novel ring-based mid-air gesture typing system for lightweight AR glasses, leveraging an intuitive pointing and selection technique. The deep learning word prediction framework, \textit{Score Fusion}, significantly enhances text entry accuracy and speed. 
User studies demonstrate \textit{RingGesture}'s effectiveness and usability, with entry rates approaching 30 WPM, outperforming previous one-handed text entry techniques for AR/VR. 
Thus, \textit{RingGesture} demonstrates significant potential for enabling fast text entry in lightweight AR glasses.

\bibliographystyle{abbrv-doi-hyperref}

\bibliography{template}

\appendix % You can use the `hideappendix` class option to skip everything after \appendix

\section{About Appendices}
Refer to \cref{sec:appendices_inst} for instructions regarding appendices.

\section{Troubleshooting}
\label{appendix:troubleshooting}

\subsection{ifpdf error}

If you receive compilation errors along the lines of \texttt{Package ifpdf Error: Name clash, \textbackslash ifpdf is already defined} then please add a new line \verb|\let\ifpdf\relax| right after the \verb|\documentclass[journal]{vgtc}| call.
Note that your error is due to packages you use that define \verb|\ifpdf| which is obsolete (the result is that \verb|\ifpdf| is defined twice); these packages should be changed to use \verb|ifpdf| package instead.

\subsection{\texttt{pdfendlink} error}

Occasionally (for some \LaTeX\ distributions) this hyper-linked bib\TeX\ style may lead to \textbf{compilation errors} (\texttt{pdfendlink ended up in different nesting level ...}) if a reference entry is broken across two pages (due to a bug in \verb|hyperref|).
In this case, make sure you have the latest version of the \verb|hyperref| package (i.e.\ update your \LaTeX\ installation/packages) or, alternatively, revert back to \verb|\bibliographystyle{abbrv-doi}| (at the expense of removing hyperlinks from the bibliography) and try \verb|\bibliographystyle{abbrv-doi-hyperref}| again after some more editing.

\end{document}